\documentclass[sigconf,screen,leftjustify]{acmart}

\AtBeginDocument{%
  \providecommand\BibTex{{%
    \normalfont B\kern-0.5em{\scshape i\kern-0.25em b}\kern-0.8em\Tex}}}

\setcopyright{rightsretained}%{acmcopyright}
\copyrightyear{2019}
\acmYear{2019}
\acmDOI{}%10.1145/1122445.1122456}

\usepackage{fancyhdr}
\acmConference[KDD '19: TMEDSC]{Knowledge Data Discovery: Tensor Methods for Emerging Data Science Challenges}{August 4-8, 2019}{Anchorage, AK, USA}
\acmBooktitle{The
25th ACM SIGKDD Conference on Knowledge Discovery and Data Mining (KDD'19): Tensor Methods for Emerging Data Science Challenges, August 04-08, 2019, Anchorage, AK}
\acmISBN{978-1-4503-6201-6/19/08}

\usepackage{times}
\usepackage{epsfig}
\usepackage{graphicx}
\usepackage{amsmath,bm}
\usepackage{amssymb}
\usepackage{algorithm}
\usepackage{appendix}
\usepackage{float}
\usepackage{caption}
\usepackage{times,amsmath,epsfig,graphicx,amssymb,multirow,subfig}
\usepackage{mathtools, cuted}
\usepackage{empheq}
\usepackage[hang,flushmargin]{footmisc} 

\usepackage{geometry}
\usepackage{geometry}
\usepackage{url}

\renewcommand{\textrightarrow}{$\rightarrow$}

\makeatletter
    \renewcommand\@makefntext[1]{\leftskip=0em\hskip-0em\@makefnmark#1}
\makeatother

\def\ie{\textit{i}.\textit{e}., }%{\emph{i.e}\onedot} 
\def\etal{\textit{et al.}}%{\emph{et al}\onedot}

\newcommand{\rpm}{\raisebox{.2ex}{$\scriptstyle\pm$}}

\newcommand\Tstrut{\rule{0pt}{2.6ex}}       % ''top'' strut
\newtheorem{definition}{Definition}

\def\Reals{\mathbb{R}}
\def\Complex{\mathbb{C}}

\def\th{^{\hbox{\small th}}}
\def\nd{^{\hbox{\small nd}}}
\def\rd{^{\hbox{\small rd}}}

\def\vec#1{{\mathbf{#1}}}
\def\mat#1{{\textbf{#1}}}

\def\ten#1{{\mathbf{\mathcal#1}}}

\def\tp{^{\mathrm{T}}}

\def\oast{\circledast}

\def\assign{\mathrel{\mathop:}=}

\def\mode#1{_{\mbox{\tiny\textrm{#1}}}}
\def\tmode#1{_{\mbox{\tiny$\ten{#1}$}}}

\def\measure{\mode 0}
\def\pixels{\mode x}

\def\people{\mode P}
\def\views{\mode V}
\def\illums{\mode L}
\def\expres{\mode E}

\def\matize#1#2{{\mat#1}\mode{$[#2]$}}

\def\timesT{\times^{\hskip-1pt\hbox{\tiny\textrm{}T}}}

\def\pinv#1{^{+\lower2pt\hbox{\hskip-2pt\hbox{\tiny${}#1$}}}}
\def\inv{^{-1}}

\renewcommand{\baselinestretch}{.9535}
\begin{document}

%% The ''title'' command has an optional parameter,
%% allowing the author to define a ''short title'' to be used in page headers.
\title[Compositional Hierarchical Tensor Factorization]{
%Generalized Block Tensor Decomposition \\
Compositional Hierarchical Tensor Factorization:\\
Representing Hierarchical Intrinsic and  Extrinsic Causal Factors
%Generalized Multilinear Block Tensor Decomposition:
%for \\Intrinsic and Extrinsic Hierachical Causal Factor Representation
}

%%
%% The ''author'' command and its associated commands are used to define
%% ''authornote'' and ''authornotemark'' commands

\author{\vspace{-.175in}M. Alex O. Vasilescu$^{2,1}$}
\email{maov@cs.ucla.edu}
\affiliation{
}

\author{\vspace{-.175in}Eric Kim$^{1,2}$}
\email{ekim@cs.ucla.edu}
\affiliation{
}

%  \institution{Institute for Clarity in Documentation}
%  \streetaddress{P.O. Box 1212}

\author{ }
\affiliation{
}

\author{ }
\affiliation{
}

\author{ }
\affiliation{
}

\author{ }  
\affiliation{ 
  \institution{$^1$\hspace{-.025in}
  Department of Computer Science}
  \institution{University of California, Los Angeles}
  %\city{Los Angeles}
 % \state{CA}
  %\postalcode{90095}
  %\country{USA}
}

\author{ }
\affiliation{\hspace{-.4in}
  \institution{$^2$\hspace{-.025in}
  Tensor Vision Technologies} 
  \city{  \hspace{-.4in}
Los Angeles}
  \state{CA}
  %\postalcode{90049}
  \country{USA}
  }

\author{ }  
\affiliation{ 
%  \institution{$^6$Pintrest}
%  \city{San Francisco}
%  \state{CA}
  %\postalcode{90095}
%  \country{USA}
}

\renewcommand{\shortauthors}{M.A.O. Vasilescu and E. Kim}

%\vspace{-.15in}
%\vspace{-.15in}
\begin{abstract}

Visual objects are composed of a recursive hierarchy of perceptual wholes and parts, whose properties, such as shape, reflectance, and color, constitute a hierarchy of
%are 
intrinsic causal factors of object appearance. However, object appearance is the compositional consequence of both an object's intrinsic and 
extrinsic causal factors, where the extrinsic causal factors are related to illumination%(\ie the location and types of light sources)
, and imaging conditions.%, all of which hinder recognition.%(\ie viewpoint, viewing direction, camera characteristics)
%, in addition to to the object's intrinsic causal factors of object appearance. %This paper introduces the concepts of intrinsic and extrinsic causation and proposes  compositional hierarchical tensor factorization that explicit represents them. 
%Therefore, this paper proposes a new compositional hierarchical tensor factorization paradigm that subsumes block tensor decomposition as a special case, % and it is applicable for object image representation, recognition and synthesis when there is minimal training data. %, 
%and demonstrates its suitability for representing and disentangling the %extrinsic multifactor causal structure and the intrinsic 
%hierarchical causal structure of object image formation.
 %intrinsic and extrinsic causal factors of object image appearance.
Therefore, this paper proposes a unified tensor model of wholes and parts, and 
introduces a compositional hierarchical %multilinear %is applicable to CNNs. 
tensor factorization % paradigm that %subsumes multilinear block tensor decomposition as a special case %that represents  
%that 
%and is suitable  for interpretable part-based object representation. %and recognition. 
%Our compositional hierarchical factorization 
that disentangles the %extrinsic multifactor causal structure and the intrinsic 
hierarchical causal structure of object image formation, and subsumes multilinear block tensor decomposition as a special case. %that represents  
% approach represents and disentangles a hierarchical set of intrinsic and extrinsic of causal factors.
%The factorization 
%computes a hierarchy of object features and 
%represents an object as a compositional interpretable %recursive %of wholes and parts 
%computes an interpretable object representation.
%representation 
%that is 
%by computing statistically invariant information to all other causal factors that impact image appearance.%, but hinder recognition, 
%such as 
%that generally hinder recognition and related to 
%%scene structure,
%illumination (i.e. the location and types of light sources), and imaging (i.e. viewpoint, viewing direction, camera characteristics).
%that models an imaged object by a single unified tensor model of wholes and %parts.
%part-based architecture gives rise to a 
The resulting object representation is an interpretable combinatorial choice of
wholes' and parts' representations %. The combinatorial choice of part-based representation 
that renders object recognition robust to occlusion 
%robust to occlusion and 
%is robust to occlusion  as a result of an object representation which is in essence a 
%Track changes is on
%30
% combinatorial choice wholes' and parts' representations, 
%bypassing 
and reduces training data
requirements. 
%Furthermore, our new tensor factorization computes an interpretable feature hierarchy that is applicable to CNNs and their variants. 
%CNNs have been shown to be an equivalent to multilinear (Tucker) tensor factorization\cite{cohen16}, but %implementation  differences  between CNNs  and  a  tensor  algebraic  approach  impacts  
%with different computational complexity, memory, and data needs  due to implementation differences between neural network and tensor factorizat ion methods.  
We demonstrate our approach in the context of face recognition
by training on an extremely reduced dataset of synthetic images, and 
report encouraging face verification results on two datasets -- the Freiburg dataset, and 
the Labeled Face in the Wild (LFW) dataset consisting of real-world images, thus, substantiating the suitability of our approach for data starved domains.%in scenarios where minimal training data exists, 

\end{abstract}

\vspace{-.0in}

\maketitle
%\newpage
%\vskip -.2in
\section{Introduction}%: Motivation and Related Work}
\label{sec:Introduction}
\vskip -.025in
%\iffalse
Statistical data analysis that disentangles the causal factors of data formation and computes a representation that facilitates the analysis, visualization, compression, approximation, and/or interpretation of the data is challenging and of paramount  importance. 
%\fi
\iffalse
Data, such as images, sound, motion data, are formed from the interaction of multiple constituent factors associated with %related to 
structure, behavioral processes, or biological processes that cause predictable patterns of variation in observed data, and are the causal factors of data formation. 
\fi
%These causal factors are related to scene structure, behavioral processes, or biological processes.% that cause patterns of variation in observed data, and are the causal factors of data formation.% that are comprised of a set of measurements. 
%Multilinear tensor data analysis was first employed in computer vision~\cite{Vasilescu01b,Vasilescu02}, computer graphics~\cite{Vasilescu04}, and machine learning~\cite{Vasilescu05} to image data in order to disentangle causal factors related to illumination (i.e. the location and types of light sources), imaging (i.e. viewpoint, viewing direction, lens type and other camera characteristics), and scene structure.
%, but without an explicit representation of individual wholes and parts of scene structure.  
%In this paper, we refine o causal factors as either intrinsic or extrinsic causal factors of object appearance.

%\iffalse
%\vspace{+.1in}
%\begin{quote}
``Natural images are the composite consequence of multiple constituent factors related to scene structure, %illumination, and imaging.  %Scene structure is determined by a collection of visual objects.
 illumination conditions, and imaging conditions. 
Multilinear algebra, the algebra of higher-order tensors, offers a potent mathematical framework for analyzing the multifactor structure of image ensembles and for addressing the difficult problem of disentangling the constituent factors or modes.''
(Vasilescu and Terzopoulos, $2002$~\cite{vasilescu02})
%\end{quote}
%\vspace{-.05in}
%\noindent
%\fi

Scene structure is composed from a set of objects that appear to be formed from a recursive hierarchy of perceptual wholes and parts whose properties, such as shape, reflectance, and color, 
%determine an object's appearance. Appearance-based properties associated with an object's recursive hierarchy %of parts of wholes and parts 
constitute a hierarchy of intrinsic causal factors of object appearance.
%While all causal factors of image formation are extrinsic to image appearance, 
%However, 
Object appearance is the compositional consequence of both an object's intrinsic causal factors, and extrinsic causal factors with the latter related to 
illumination (i.e. the location and types of light sources),
  imaging (i.e. viewpoint, viewing direction, lens type and other camera characteristics).  
%The meaning and nature of i
 Intrinsic and extrinsic causal factors confound each other's contributions hindering recognition.

%\begin{figure}[!t]
%  \includegraphics[width=.5\textwidth]{images/3layers_generic_compositional_example14_I0_3.png}\\
%  \caption[width=.5\textwidth]{Seattle Mariners at Spring Training, 2010.}
%  \caption{Representing a vectorized image and a data tensor, $\ten D$, in terms of a hierarchical data tensor, $\ten D\tmode H$.}
%    \vspace{-.15in}
%\end{figure}

``Intrinsic properties are by virtue of the thing itself and nothing else'' (David Lewis, 1983~\cite{Lewis1983a}); whereas an extrinsic properties are not entirely about that thing, but as result of the way the thing interacts with the world.  
 % If something has an intrinsic property, then when situated in different surroundings will differ in its % extrinsic properties. (Lewis 1983: 197)  
%~\footnote{
%David Lewis~\cite{Lewis1983a} provides formal discussion of intrinsic and extrinsic concepts of causality and addresses a few related distinctions that an intuitive definition conflates, such as local versus global intrinsic properties, duplication preserving properties, and interior versus exterior properties.  
%The meaning and nature of intrinsic and extrinsic causation is a topic extensively explored in 
%in philosophy, ethics,philosophy of mind, metaphysics and philosophy of physics~\cite{Hume1748,Lewis1973,Lewis1983a,Menzies1999}.  In this paper, a hierarchy of intrinsic causal factors are represented as statistical invariant information of other causal factors. 
%The distinction between intrinsic and extrinsic properties is intuitive enough, however an intuitive understanding conflates a few characterizations related to locality, duplication preservation and several kinds of interiority.}
%Most important to this pap er is the definition of local and global intrinsic properties. 
Unlike global intrinsic properties, local intrinsic properties are intrinsic to a part of the thing, and it may be said that a local intrinsic property is in an ``intrinsic fashion'', or ``intrinsically'' about the thing, rather than ``is intrinsic'' to the thing~\cite{Humberstone1996,Figdor2008}. 
David Lewis~\cite{Lewis1983a} provides a formal discussion of intrinsic and extrinsic concepts of causality and addresses a few related distinctions that an intuitive definition conflates, such as local versus global intrinsic properties, duplication preserving properties, and interior versus exterior properties. 
The meaning of intrinsic and extrinsic causation was extensively explored in philosophy, %ethics, 
philosophy of mind, metaphysics 
and philosophy of physics~\cite{Hume1748,Lewis1973,Lewis1983a,Menzies1999,pearl00}.
%We define the meaning of intrinsic and extrinsic causation consistently with the nature and meaning explored in philosophy, ethics, philosophy of mind, metaphysics and philosophy of physics~\cite{Hume1748,Lewis1973,Lewis1983a,Menzies1999}.% and we follow their definition. 

Our goal is to explicitly
represent local and global 
%hierarchical 
intrinsic causal factors as statistically invariant representations to all other causal factors of data formation.%~\cite{rubin77,pearl00}.

Historically, 
statistical object recognition paradigms
%parallel human perception paradigms and 
can be categorized 
based on how object structure is represented and recognized, ie. based
on the appearance of an object's local features~\cite{Wong2012,Gao2010}, or based on the overall global object appearance~\cite{Heisele2003,Shakhnarovich04,Bartlett02, Yang2001,Martinez2001,belhumeur1997,Turk91b}. Both approaches have strengths and shortcomings.  Global features are sensitive to occlusions, while local features are sensitive to local deformations and noise.  A hybrid approach that employs both global object features, and local features mitigates the shortcoming of both approaches~\cite{Li2015,Murphy2006,xu08}. 

Deep learning methods, which have become a highly successful approach for object recognition, compute feature hierarchies composed of low-level and mid-level features either in a supervised~\cite{lecun98}, unsupervised~\cite{hinton06,larochelle07,ranzato06} or semi-supervised manner~\cite{nair09}. This has been achieved by composing modules of the same architectures, such as Restricted Boltzmann Machines~\cite{hinton06}, autoencoders%\footnote{Autoencoders implement PCA(SVD) in a neural network framework.}
~\cite{larochelle07}, or various forms of encoder-decoder networks~\cite{ranzato06,Cadieu09,Jarrett09}.
%Models with excessive number of parameters have increased
%storage and computation requirements rendering them null-and-void for deployment on devices with
%limited resources.

%Given that autoencoders are inefficient neural network implementations of PCA, a method based on linear algebra that employs the matrix SVD, it should not be surprising that t

\begin{figure*}[!tbph]
%\vspace{-.05in}
\begin{tabular}{cc}
%\vskip+.25in
\hskip-.25in
\begin{tabular}{c}
\vspace{-.3in}
%\vskip+.2in
\\
\\
%includegraphics[width=1.1\linewidth]{images/Block-Tucker-Compressed-Generic.png}
\psfig{figure=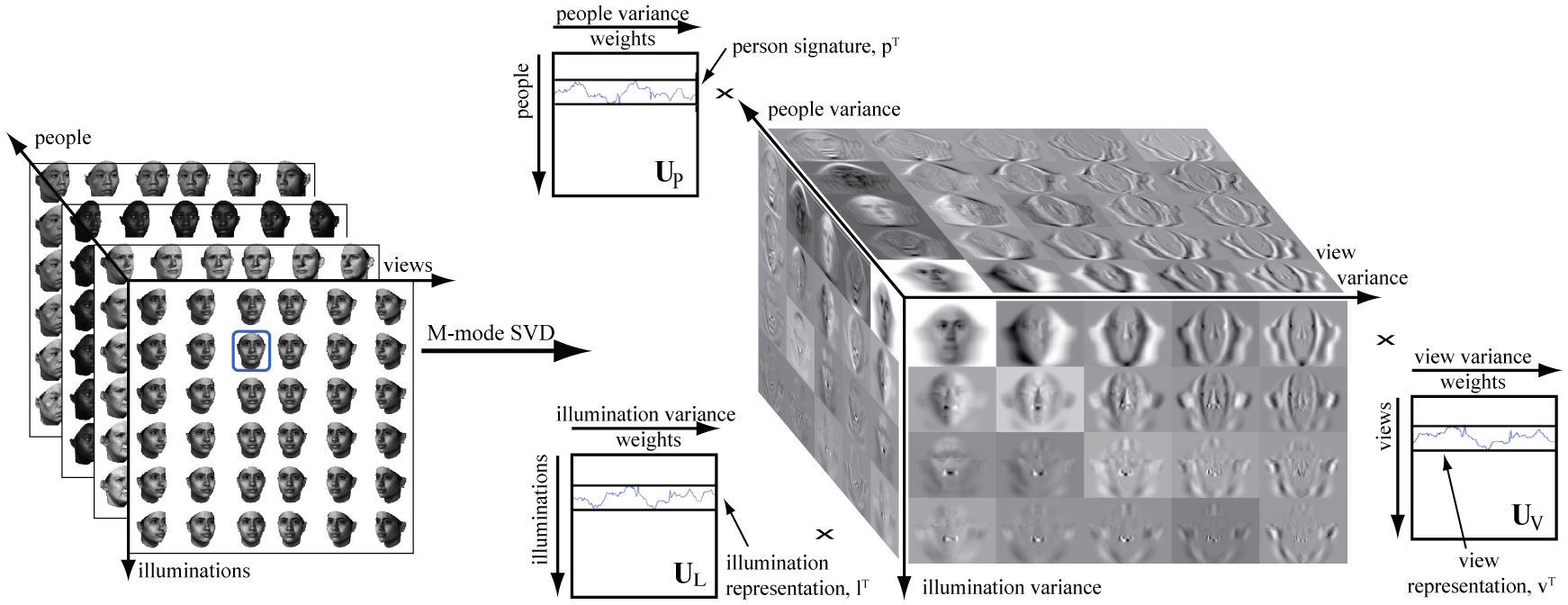,width=.65\linewidth}\\
%\vskip+.1in
\vspace{-.225in}\\
\mbox{
\scriptsize{(a)}}
%\centerline{\hfill\mbox{(a)}\hfill}
\end{tabular}
%\vskip-1.5in

\hskip-.4in
\vspace{-.2in}
\begin{tabular}{c}
\vspace{-.05in}
%\vskip-1.5in
%\centerline{
%\hskip+1.5in\hfill\mbox{No Cast Shadow}\hfill\\
\hskip+1in\mbox{\hskip+.1in\scriptsize{}No}\hskip+.5in\mbox{\scriptsize{}No Cast Shadow}\hskip+.2in\\
\vspace{-.05in}
\hskip+.15in\mbox{\hskip+.15in\scriptsize{Original}}\hskip+.3in\mbox{\scriptsize{}Cast Shadow}\hskip+.4in\mbox{\scriptsize{}No Shading}\\
%\centerline{

%\centerline{
\hskip+.25in%\hfill
\psfig{figure=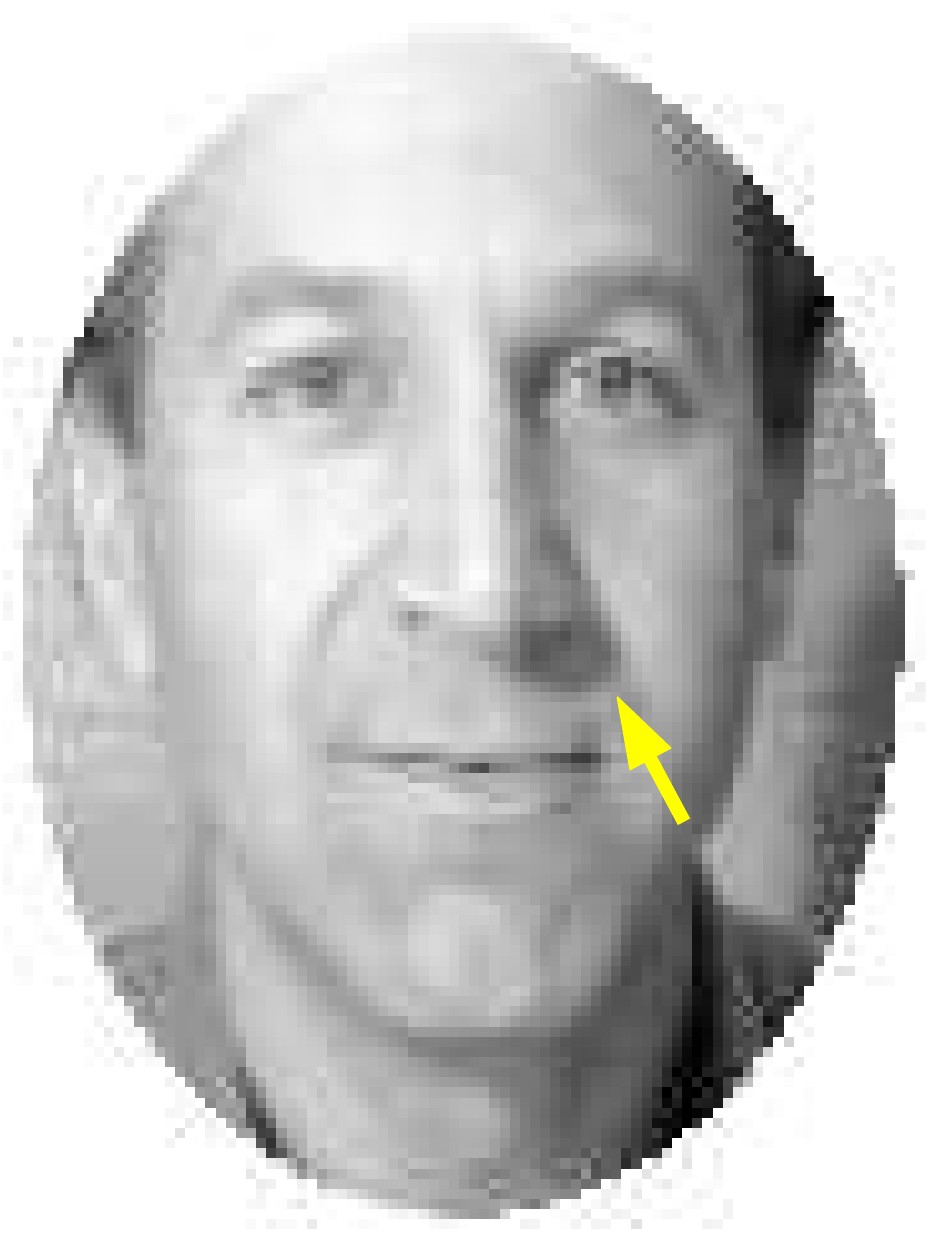,width=.6in} %\hskip 1cm
\hskip+.05in%\hfill
\psfig{figure=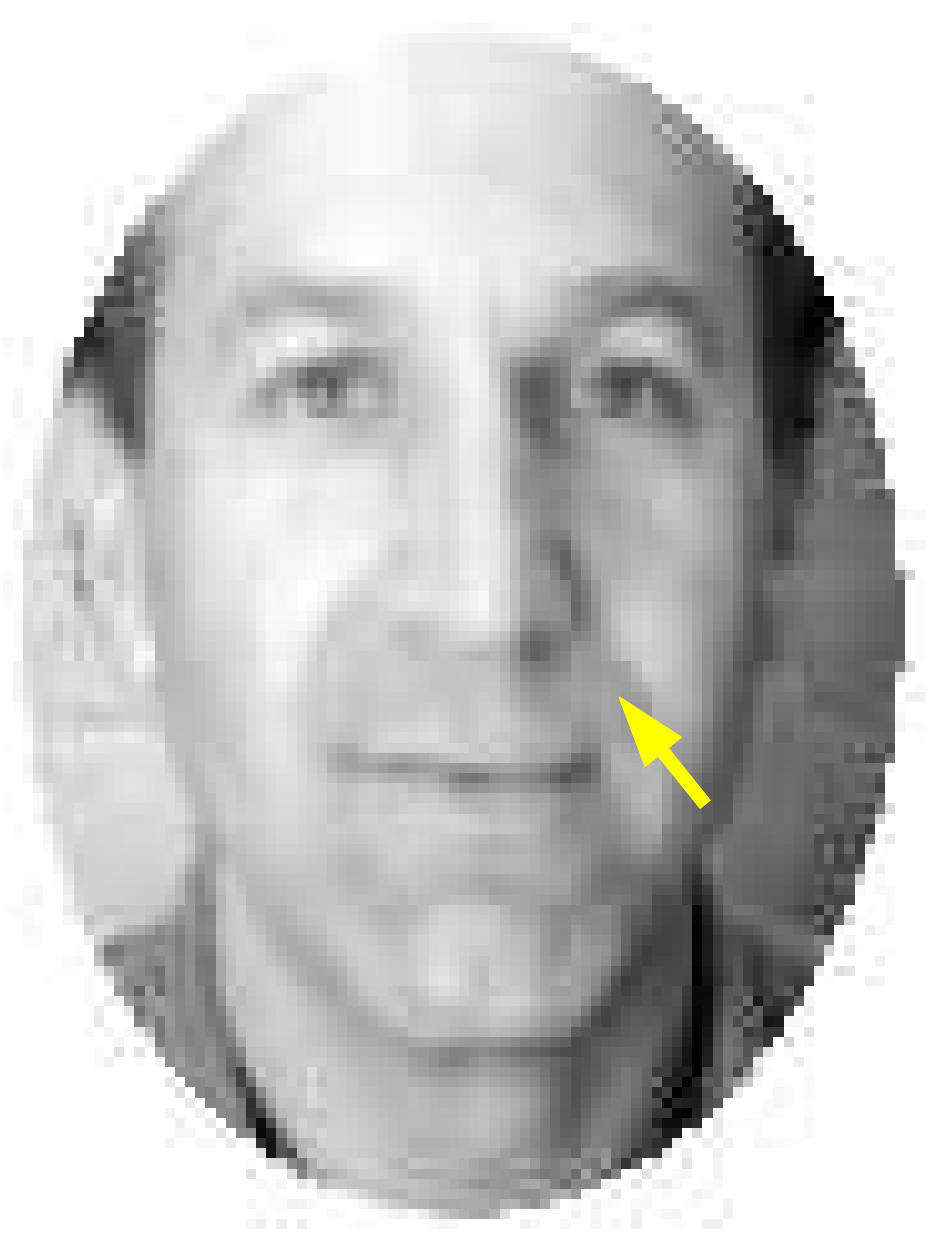,width=.6in}
\hskip+.15in
\psfig{figure=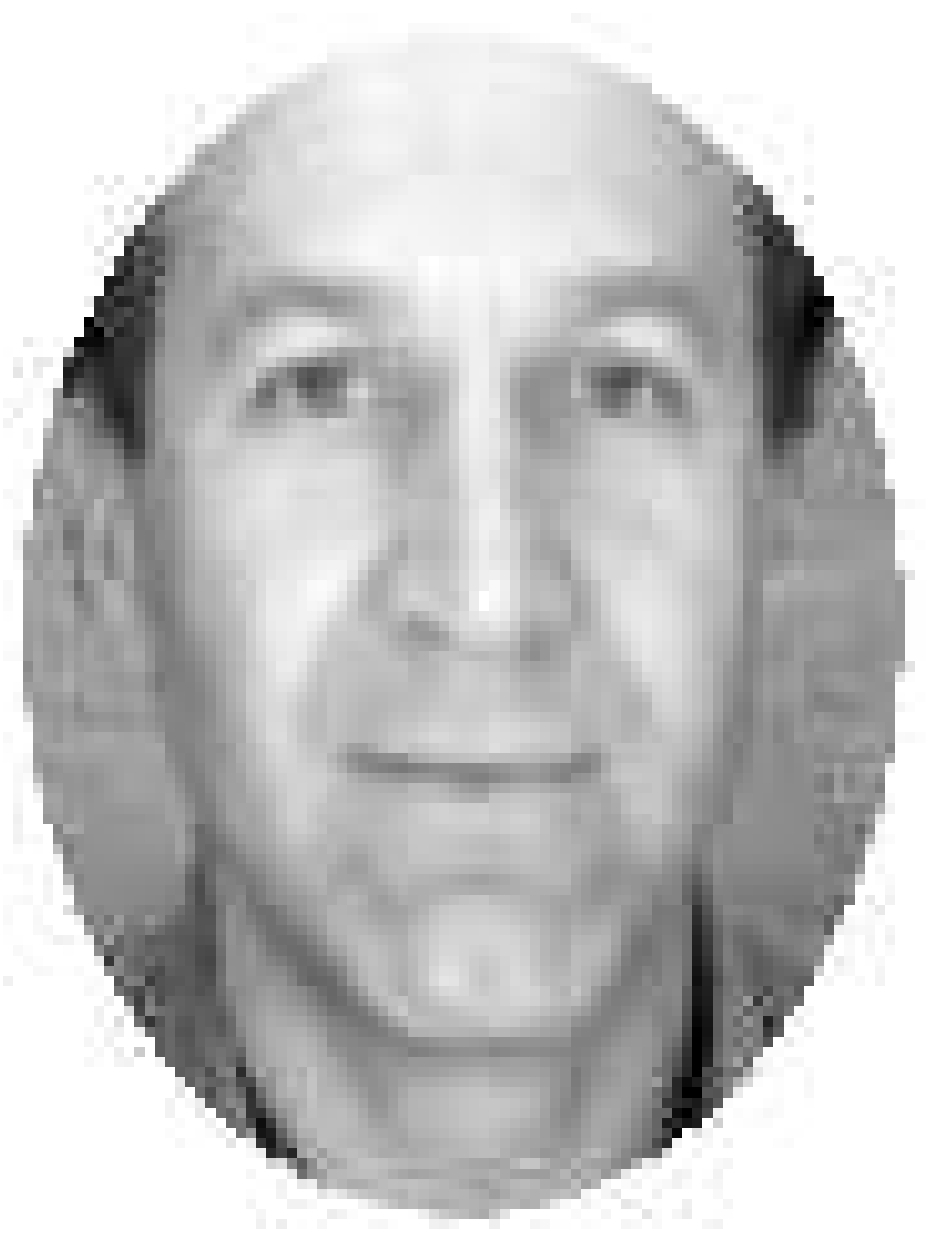,width=.6in} %\hskip 0.75cm
\hfill
\\
\vspace{-.25in}
\\
\hskip+.1in\mbox{\scriptsize{}(b)}\\
\hskip+.4in
\psfig{figure=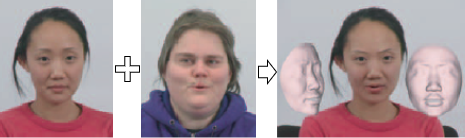,width=2.25in}\\
\vspace{-.2in}
\\
\hskip+.1in\mbox{\scriptsize{(c)}}
\end{tabular}\\
\end{tabular}
%\centerline{\hskip+3in\mbox{(a)}\hskip+3in\mbox{(c))}\hskip+1.1in}
%\centerline{(a)}
%\bigskip\bigskip
%\hskip
%-1cm\centerline{\psfig{figure=mpca_contrib_tensor6c.pdf,width=0.8\linewidth}}
%\centerline{(b)}
%\centerline{\hfill \makebox[0.4\linewidth]{(a)} \hfill 
%\makebox[0.4\linewidth]{(b)} \hfill}
%\smallskip
\vspace{+.025in}
\caption[MPCA image representation and coefficient vectors]{(a) TensorFaces representation: This image illustrates a multilinear tensor factorization of a $4^{th}$-order training data tensor, $\ten D\in\Reals^{I\pixels \times I\illums \times I\views \times I\people}$ where $I\pixels, I\illums, I\views, I\people$ are the number of pixels, illuminations, views and people, respectively. %The data tensor, $\ten D$ is formed by a set of images synthesized in Maya from  3D scans of 100 subjects($I\people=100$), recorded using a Cyberware${}^\mathrm{TM}$ 3030PS laser scanner as part of the University of Freiburg 3D morphable faces database ~\cite{Blanz99}. 
A vectorized image, $\vec d$, is represented by a set of coefficient vectors, one for the illumination, viewpoint, and person, and expressed mathematically as $\vec d=\ten T\times\illums \vec l\tp
\times\views \vec v\tp\times\people \vec p\tp$, where 
 the TensorFaces basis, $\ten T$, governs the interaction between different causal factors of data formation. (Data should be centered, but for this display it was added back.)
%A portion of the $4\th$-order data tensor
%$\ten D$ of the image ensemble used for training.
%formed from the dash-framed images of each
%person in Figure~\ref{fig:face-data}. 
%Only 4 of the 75 people are shown.
(b) These set of images demonstrate the models ability to disentangle the causal factors. Illumination effects, such as highlights, cast shadows and shading ~\cite{Vasilescu03} are predectively and progressively reduced. (c) Vlasic\ etal~\cite{Vlasic05} generate new images by performing multilinear expression re-targeting.}
\iffalse
While TensorFaces (Multilinear-PCA or Multilinear-ICA) learns the interaction between the causal factors from training data, it does not prescribe an approach for determining the causal factors from one or  more unlabeled test images.  
For an unlabeled
vectorized test image $\vec d\mode{new}$, 
%the associated coefficient vectors $\vec p$, $\vec v$, $\vec l$ 
the causal factor labels are estimated through a multilinear projection algorithm~\cite{Vasilescu2011} which decomposing the response tensor 
%(in machine learning this would be known as an activation tensor)
$\ten R=\ten T\pinv\pixels\timesT\pixels \vec d\mode{new}\approx \vec r\illums \circ \vec r\views \circ \vec r\people$  
into $\vec r\illums$, 
$\vec r\views$, and $\vec r\people$ the representations for the inferred view, illumination and person label.
\fi
\label{fig:tensorfaces}
\vspace{-.15in}
\end{figure*}

Cohen \etal's ~\cite{cohen16,cohen16b}
%  collobert2011torch7, Bastien-Theano-2012independent, bergstra+al:2010-scipy}
 %have demonstrated by employing measure theory and matrix algebra 
theoretical results show
 that convolutional neural networks (CNNs) %, neural network architectures that may be composed from a set of autoencoder modules, 
 are theoretically equivalent in their representational power to hierarchical Tucker factorization
 ~\cite{hackbusch2009,Grasedyck2010,perros15,oseledets11},%,Tucker66, Delathauwer00a}
 \iffalse
 \footnote{Hierarchical Tucker is a resource efficient approximation of the Tucker tensor factorization (also known as, the M-mode SVD, Multilinear SVD, multilinear PCA)~\cite{Tucker66, Delathauwer00a,Vasilescu02a,Vasilescu02}.  Note, that problem setup makes a difference in the outcome of applying the M-mode SVD. }
 \fi
 \footnote{While the Tucker factorization performs one matrix SVD to represent a causal factor subspace (one orthonormal mode matrix), %omputation
 %represents one of the causal factors of data formation, 
 a Hierarchical Tucker is a hierarchical computation of the Tucker factorization that employs a stack of SVDs to represent a causal factor subspace (one orthonormal mode matrix). For computational efficiency, the authors~\cite{hackbusch2009,Grasedyck2010} prescribe a stack of QR decompositions instead of a stack of SVDs.
}
  and shallow networks are equivalent to linear tensor factorizations, aka CANDECOMP/Parafac (CP) tensor factorization~\cite{Carroll70,Harshman70,Bro97}. 
%Since CNNs learn millions of parameters that may lead to redundancy and poor generalization, there has been a research thrust to reparameterize and reduce the number of CNN parameters by organizing CNNs weights into a tensor, and employing tensor factorizations~\cite{Lebedev14,Nikolov15,Kim16}. While we do not explore this research direction in this paper, our tensor factorization is applicable. 
Vasilescu and Terzopoulos~\cite{Vasilescu01b}\cite{vasilescu02}\cite{Vasilescu03}\cite{Vasilescu05}\cite{Vasilescu2011} %re-framed the analysis, recognition, synthesis, and interpretability of sensory data as multilinear tensor factorization problems, noting 
demonstrated that tensor algebra is a suitable,\footnote{
The suitability of the tensor framework was also demonstrated in the context of computer graphics by synthesizing new textures ~\cite{Vasilescu04}, performing expression retargeting/ reanimation~\cite{Vlasic05} %,Macedo06}
and by synthesizing human motion~\cite{Vasilescu01a}\cite{Vasilescu02a}\cite{Hsu05}.}
%representations of data formation, as well as generating new data. based on how the problem is set up.
 interpretable framework for mathematically representing and disentangling the causal structure of data formation in computer vision, computer graphics and machine learning
%~\footnote{There are
%second-order and %higher-order %statistical methods %that compute accurate, %or statistically %meaningful
%representations of the %causal factors, as well %as kernel %variants
~\cite{Vasilescu05,Vasilescu09},
%}
%demonstrated that the tensor framework is suitable for explicitly representing and %demonstratively disentangle the causal structure of data formation
 Figure~\ref{fig:tensorfaces}. %, and 
%and the first step and seemingly unimportant one requires that data is vectorized. represents and disentanglesthe hierarchical causal structure of object image formatio
%whose tensor factorization 
%an interpretable approach, and a major shortcoming of current CNNs. 
Problem setup and implementation differences %\footnote{Our multilinear tensor factorization approach is a supervised method that treats an image as a vector, computes statistics between all pixel combinations in an image, nearby and far away pixels and determines invariant representations for all causal factors. Cohen etal, has shown an equivalence between CNNs and a multilinear tensor factorization approach that is unsupervised, and treats an image as a matrix. In this case, pixels statistics are computed only for combination of pixels that are co-located in the same column or the same row, ignoring all diagonal pixels. The later approach also does not compute causal factor representations.} 
 between CNNs and our tensor algebraic approach 
impact interpretability, data needs, memory/storage and computational complexity, often rendering CNN models %require millions of images to train and storage for millions of parameters
%storage and computation requirements 
%rendering them m
difficult to deploy on mobile devices, or any devices with limited computational resources.
%CNNs require millions of training images~\cite{Taigman14,Russakovsky15} and learn millions of parameters that can lead to redundancy and poor generalization.
%Thus, tensor methods have become a powerful approach for learning various causal factors (latent variables) of data formation.   

Inspired and inspirited by Cohen~\etal's~\cite{cohen16,cohen16b} theoretical results, 
%who have demonstrated that convolutional neural networks (CNN) are equivalent to a Hierarchical Tucker (multilinear) tensor factorization~\cite{hackbusch2009,Grasedyck2010,Tucker66, Delathauwer00a}\footnote{Hierarchical Tucker is a resource efficient approximation of the multilinear (Tucker) tensor factorization ~\cite{Tucker66, Delathauwer00a,Vasilescu02a,Vasilescu03}.}
and %inspirited 
by the TensorFaces and Human Motion Signatures approach
~\cite{vasilescu02,Vasilescu03,Vasilescu01b,Vasilescu02b},
%%that explicitly represent, and
%demonstratively disentangle~\cite{Vasilescu03}
%global intrinsic and extrinsic causal factors of data formation, thus, enabling synthesizing new data through recomobination or retargetting~\cite{Vlasic05}, %,
%Recognizing the value of tensor factorization, and 
%Recognizing the value of both local and global features for object representation, 
%it is tempting to put forth a hybrid solution that is equivalent to in which Tucker tensor factorization of parts and sub-part and an object is represented as a concatenation of part representations.  
 we propose a unified tensor model of wholes and parts based on a reconceptualization of the data tensor as a  hierarchical data tensor, a mathematical representation of a tree data-structure.
 %\footnote{For the last 15 years plus, there have been several authors that have argued that an image should be treated as a matrix or a tensor in order to take advantage of local statistical information associated with object parts. This is a mathematically provable false statement~\cite{Vasilescu09}, but the approach has lead to a lot of papers, citations, some complex mathematics that always leaves the reader feeling smart for understanding the work.  However, those were papers solving problems of the authors own making. Our approach is embarrassingly simple and intuitive in retrospect that will leave the reader feeling mathematically unchallenged and unsatisfied. Do not be taken in by the simplicity of our solution.}
%that re-framing the engineering solution into 
%As discussed above many people have performed a hybrid part-based object recognition using various models eveloping the framework for representing an object hierarchy in a tensor framework is novel and important. 
Defining a hierarchical data tensor 
%A mathematical representation of a tree data-structure 
enables a single elegant mathematical model that can be optimized in a principled manner instead of employing a myriad of %part-based 
individual part-based engineering solutions that independently represent each part and attempt to compute all possible dimensionality reduction permutations. %To that end, we
%and formally address the intrinsic and extrinsic causal factors that may be computed by a new compositional hierarchical tensor factorization paradigm for modeling the interaction between a hierarchy of intrinsic causal factors and extrinsic causal factors of data formation.
 Our %compositional hierarchical tensor 
 factorization optimizes simultaneously across all the wholes and parts of the hierarchy, %computes the recursive part-based representation.% either with a single computation, or in a hierarchical manner where the parent whole representation is based on the child representation computation.  
learns a convolutional feature hierarchy of low-level, mid-level and high-level features, and computes an interpretable compositional object representation of parts and wholes. %, while factoring out the causal factors that hinder recognition.  
Our resulting object representation is a combinatorial choice of part representations, that renders object recognition robust to occlusion while bypassing large training data requirements.
\iffalse
The resulting object representation is a combinatorial choice of part representations which renders recognition robust to occlusion, and reduces large training data requirements.
\fi
\iffalse
, such as illumination, imaging, etc. %high level, mid-level and low level features 
Our tensor model learns a convolutional feature hierarchy of low-level and high-level features which is employed in computing a hierarchical compositional object representation invariant of extrinsic causal factors.
We demonstrate our factorization approach in the context of facial verification problem.
Hierarchical Tucker tensor factorization is an approach to computing and representing the standard Tucker tensor factorization~\cite{Tucker66}, also known as multilinear-SVD~\cite{Delathauwer00a}, that is memory and time efficient. 
Tensor factorizations have been employed for computer graphics, computer vision and machine learning since the early 2000s ~\cite{Vasilescu02a,vasilescu03,Vasilescu04,Vasilescu05,vasilescu2011} and would benefit by being implemented using a Hierarchical Tucker.
Regardless of implementation, the approaches factorize observed data and explicitly represent the causal factors of data formation. However, object representations are not compositional representations of parts and wholes.
\fi

Our compositional tensor factorization and learnt feature hierarchy is also applicable to CNNs. % and their variants.
%Furthermore, s
Since CNNs learn millions of parameters that may lead to redundancy and poor generalization,  
%Our factorization may be used to reparameterize and reduce the number of CNN parameters 
%Organizing CNN parameters into a tensor and performing tensor dimensionality 
%reduction~\cite{Lebedev14,Novikov2015,Kim15,kossaifi2017tensor}, 
our factorization and dimensionality reduction approach may be employed to reparameterize and reduce a tensor of CNN parameters, potentially resulting in better generalization~\cite{Lebedev14,Novikov2015,Kim15,kossaifi17}.
%Our compositional hierarchical tensor factorization is applicable in this scenario as well.
%While we do not explore this research direction in this paper, our tensor factorization is applicable. 
%models with excessive number of parameters have increased
%storage and computation requirements rendering them null-and-void for deployment on devices with
%limited resources.

DeepFace, a CNN approach~\cite{Taigman2014,Huang2012,Sun2013,Chen2015,xiong2016},
celebrated closing the gap on human-level performance for face
verification by testing on the Labeled Faces in the Wild (LFW) database~\cite{Huang2007} of $13,233$ images from $5,749$ people in the
news, and training on a large dataset of $4,400,000$ facial
images %of $200\times200$ pixels by collecting $\sim\hskip-.05in1000$ images
from $4,030$ people, the same order of magnitude as the number of people in the test data set ~\cite{Parkhi15}.  %Not suprisingly, t
 This large training data resulted in an increase from $70\%$ verification rates to $97.35\%$.  %This result should be considered an abject failure.

The very resources that make deep learning a wildly successful approach today are also its shortcomings.  In general, it is difficult and expensive to acquire large
representative training data for object image analysis or recognition,
and once acquired, there is a need for high performance computing, such as distributed GPU computing
~\cite{tensorflow2015-whitepaper, collobert2011torch7,
  Bastien-Theano-2012, bergstra+al:2010-scipy}. %The deployment of these types of models on devices with limited %resources null-and-void in their present form.
% and result in multiple representations per person. In the limit, there
 % can be as many representations as there are images rather than a
% unique signature per person, making real-time classification
% challenging.

%Ideally, one ought to sample the image manifold at twice the Niquist rate.

%and are resorting to adhoc 
%Models with excessive number of parameters have increased
%storage and computation requirements rendering them null-and-void for deployment on devices with
%limited resources.

While we have not closed the gap on human performance, the expressive power of our representation and our verification results are promising.  We demonstrate and validate our novel compositional tensor representation in the context of the face verification, albeit it is intended for any data verification or classification scenario.
%We demonstrate our new compositional tensor representation in the context of the face verification, however, it is intended for any type of data verification or classification.% rather than face verification per se.  
We trained on less than one percent ($1\%$) of the total images used by DeepFace. We trained on the images of only $100$ people and tested on the images of the $5,749$ people in the LFW database. 
%approximately half of one percent 
%of the total images used by DeepFace. 
%Images were synthetically generated from 3D scans of the 100 subjects~\cite{Blanz99}.
% (see Supplemental Material), 

%recorded using a Cyberware${}^\mathrm{TM}$ 3030PS laser
%scanner as part of the University of Freiburg 3D morphable faces
%database ~\cite{Blanz99}.

\iffalse
This paper introduce
(i) the concepts of intrinsic and extrinsic causal factors, 
(ii) proposes a novel unified tensor model of whole and parts, and 
(iii) introduces a new hierarchical compositional tensor factorization which either computes the representation for the object recursive hierarchy with a single factorization, or in an incremental manner where the parent whole representation is dependent on the child part representation. 

\fi

\vspace{+.05in}
\noindent
\underline{Contributions:}
\iffalse
In the last $5-10$ years, CNN research which has been financially backed by commercial interests, such as Facebook and Google, has been dominating nearly every scientific field, recruiting tens of thousands of the best minds around the world to the detriment, near exclusion, and threat of extinction of every other line of scientific inquiry.
The billion-dollar commercial backing of a single line of inquiry constitutes a monopoly of thought, a monopoly of the best minds this world has to offer which will have the same type of deleterious effects as any other type of monopoly, impacting education, thought diversity (a travesty to the scientific community), and eventually negatively impacting innovation.%~\footnote{Historically, the US government has financially backed multiple lines of inquiry in computer vision and more generally AI and divested public investment once an approach became mature enough to be of interest to industry.}
 Many of those conducting research in CNN either ignore or unaware of research in other fields often operating as scientific silos, 
 even tensor algebraic research is relatively  unexplored despite its close connection to CNNs. 
\fi
\vspace{+.05in}
\noindent
%Therefore, we find ourselves in the bizare situation of having to defend the need to extend the tensor algebraic framework.  
%In light of today's scientific environment, the contribution of this paper is three fold.
  
{$\mat 1.$} This paper (i) explicitly addresses the meaning of intrinsic versus extrinsic causality% that was extensively explored in philosophy of mind, metaphysics 
%and philosophy of physics~\cite{Hume1748,Lewis1973,Lewis1983a,Menzies1999}
, and (ii) models cause-and-effect as multilinear tensor interaction between intrinsic and extrinsic hierarchical causal factors of data formation. The causal factor representations are interpretable, hierarchical, statistically invariant to all other causal factors and computed based on $2\nd$ order statistics, but may be extended to employ higher-order 
statistics, or kernel approaches. 
%statistically invariant representations to all other causal factors computed %based on 2nd order statistics, that may be extended to employ higher-order %statistics, or kernel approaches. 

\iffalse
{\bf 1.} This paper 
formally addresses and defines the meaning of intrinsic versus extrinsic causality. We explicitly represent a hierarchy of intrinsic and extrinsic causal factors of data formation as statistically invariant representations to all other causal factors based on $2\nd$ order statistics which may be extended to employ higher order statistics, or kernel approaches. Vasilescu and Terzopoulos provide a gentle introduction of the field in their TensorFaces work~\cite{Vasilescu02a}.% and vasilescu gives 
Kolda and Bader~\cite{Kolda09} provides a more comprehensive review of the tensor algebraic approach. 
%
%This paper 
%It is an alternate implementation approach to CNNs%the neural network implementation, %It is a tensor algebraic implementation approach to CNNs that results in a deep learning model that is interpretable and employs minimal data. 
\fi

{$\mat 2.$} %This paper represents an alternate line of thought and inquiry. 
 %This paper contributes to the diversity of thought of the scientific community.  
In analogy to autoencoders which are inefficient and approximate neural network implementations
of principal component analysis, a pattern analysis method based in linear algebra, CNNs are neural network implementations of tensor factorizations. 

%{$\mat 3.$} 
%Unlike matrix algebra, tensor algebra is relatively undeveloped and t
This paper contributes to the tensor algebraic paradigm:
%framework, making the following specific technical contributions to the tensor algebraic framework: 
%This paper extends the tensor mathematical foundation and 
%efficient compositional part-based representation, 
%(i) a discussion of the concepts of intrinsic and extrinsic causality, 
%that takes advantages of the spareness and the hierarchical nature of our model to compute an efficient compositional part-based representation, 
%\\
$\mat{(i)}$ we express our data tensor in terms of a unified tensor model of wholes and parts 
by defining a hierarchical data tensor; 
%\\
$\mat{(ii)}$ %our explicit mathematical re-expression of our data tensor in terms of a hierarchical data tensor
we introduce a compositional hierarchical tensor factorization that subsumes block-tensor decomposition as a special case~\cite{DeLathauwer08,Delathauwer08c}; 
%\\
\iffalse
$\mat{(iii)}$ we introduce a hierarchical computation of %the compositional hierarchical 
our new factorization based on a new and alternate hierarchical computation of the multilinear (Tucker) tensor decomposition;
%\\
\fi
$\mat{(iii)}$ we validate our approach by employing our new compositional hierarchical tensor factorization % to represent and disentangle extrinsic multifactor structure and the intrinsic hierarchical structure of data formation
%when applying a standard multilinear factorization ($M$-mode SVD/Tucker factorization).  This is a conceptual contribution rather than the way one ought to compute a compositional part-based representation.  This is a mathematical representation of a data tensor as a unified hierarchical tensor model of whole and parts.% that reconceptualizes the data tensor as a hierarchical data tensor, 
%two alternates with different advantages and disadvantages 
%a incremental/hierarchical set of factorizations where the parent whole representation is computed based on the child part computation,
%and 
%(iv) validate of our approach
in the context of face recognition, but it may be applied to any type of data.%, such as scarce highly-dimensional medical data.%~\cite{landau19,kohli17}. %,hatamizadeh2018automatic}.
%\\
%\noindent
This approach is data agnostic.% and may be applied to non-imaging data.

\section{Relevant Tensor Algebra}
%\vspace{-.05in}
We will use standard textbook notation,denoting scalars by lower case italic letters $(\it{a, b, ...})$, vectors by bold lower case letters $(\vec{a, b, ...})$, matrices by bold uppercase letters $(\mat A, \mat B,...)$, and higher-order tensors by bold uppercase calligraphic letters $(\ten A, \ten B,...)$. Index upper bounds are denoted by italic uppercase (\ie  $1\le i \le I$).  The zero matrix is denoted by $\mat 0$, and the identity matrix is denoted by $\mat I$. 
%Appendix~\ref{sec:review} overviews the basics of tensor algebra.

\begin{figure}[!t]
\vspace{-.05in}
\centerline{\includegraphics[width=.81\linewidth]{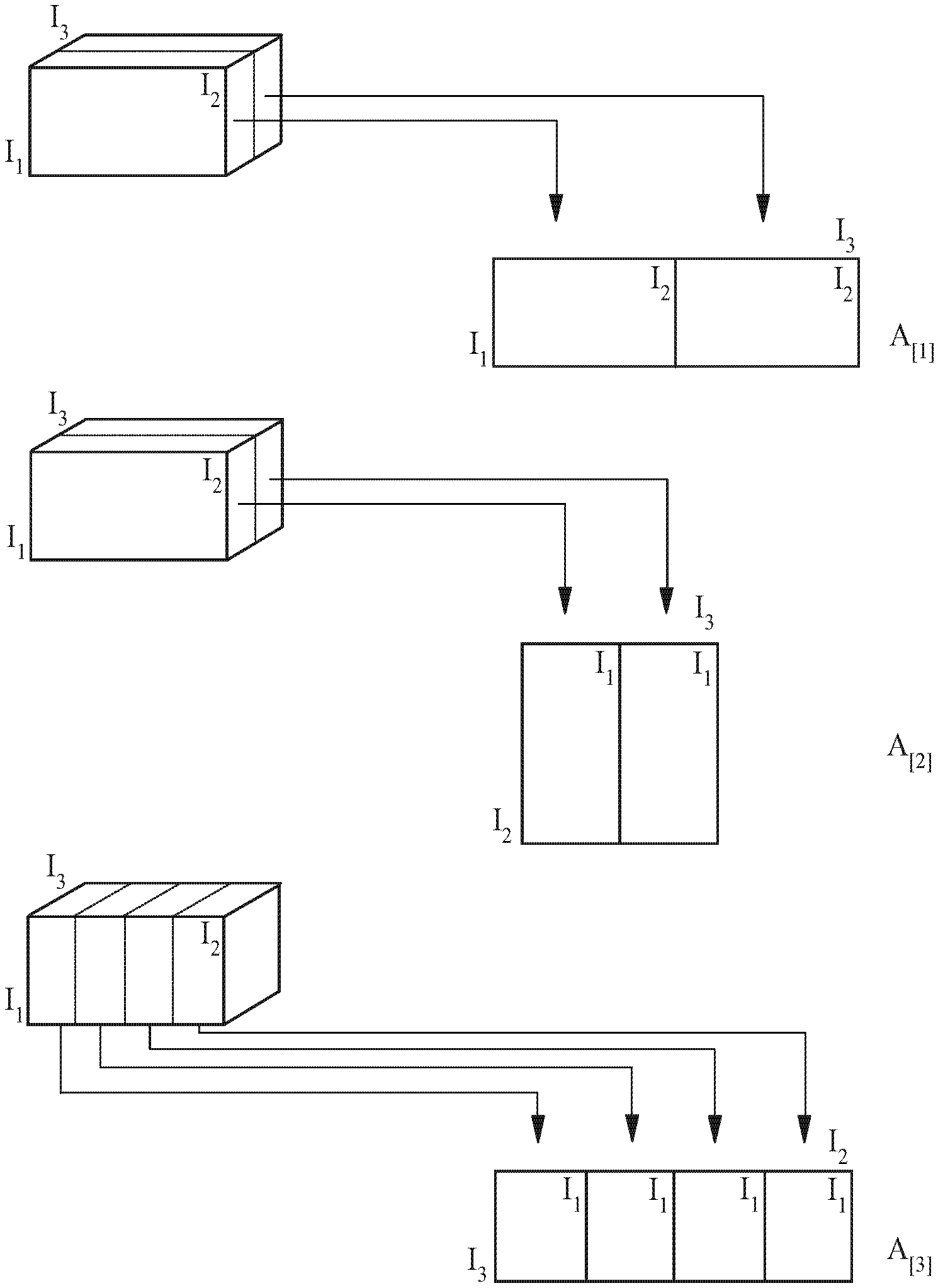}}
%\centerline{\psfig{figure=flatten3_new.png,width=3in}}
%\centerline{\psfig{figure=flatten2.pdf,width=3in}}
%\centerline{\psfig{figure=flatten.pdf,width=3in}}
%\centerline{\psfig{figure=flatten3.pdf,width=3in}}
\vspace{-.1in}
\caption[Matrixizing a (3rd-order) tensor]{Matrixizing a (3rd-order)
tensor. The tensor can be matrixized in 3 ways to obtain matrices
comprising its 1-mode, 2-mode, and 3-mode vectors. Note that this matrixizing is not cyclical, unlike the one defined in ~\cite{Delathauwer00a}.}
\label{fig:flat-tensor}
\vskip-0.45in
\end{figure}

Briefly, a {\it tensor}, or $m$-way array, is a generalization of a vector
(first-order tensor) and a matrix (second-order tensor).
%\begin{definition}[Tensor]
Tensors are multilinear mappings over a set of vector spaces. The {\it
order} of tensor $\ten A \in \Reals^{I_1 \times I_2 \times \dots
\times I_M}$ is $M$. An element of $\ten A$ is denoted as $\ten A_{i_1
\dots i_m \dots i_M}$ or $a_{i_1 \dots i_m \dots i_M}$, where $1\le
i_m \le I_m$.
%\end{definition}
In tensor terminology, column vectors are referred to as mode-1
vectors and row vectors as mode-2 vectors. The mode-$m$ vectors of an $M\th$-order tensor $\ten A\in\Reals^{I_1
\times I_2 \times \dots \times I_M}$ are the $I_m$-dimensional vectors
obtained from $\ten A$ by varying index $i_m$ while keeping the other
indices fixed. The mode-$m$ vectors of a tensor are also known as {\it fibers}. The
mode-$m$ vectors are the column vectors of matrix $\matize A m$ that
results from {\it matrixizing} (a.k.a. {\it flattening}) the tensor
$\ten A$ %(Figure 3.1 in~\cite{Vasilescu09}).%
%(Appendix~\ref{sec:review} 
(Fig.~\ref{fig:flat-tensor}).
%\vspace{-.2in}
%\iffalse
%\vspace{-0.05in}
\begin{definition}[Mode-$m$ Matrixizing]
The mode-$m$ matrixizing of tensor $\ten A \in \Reals^{I_1\times
I_2\times \dots I_M}$ is defined as the matrix $\matize A m \in
\Reals^{I_m \times (I_1 \dots I_{m-1} I_{m+1} \dots I_M)}$.
% $\matize A m \in \Reals^{I_m \times \prod_{j\neq m} I_j }$
% \Reals^{I_m \times\prod_{j=1\atop j\neq m}^M I_j
As the parenthetical ordering indicates, the mode-$m$ column vectors
are arranged by sweeping all the other mode indices through their
ranges, with smaller mode indexes varying more rapidly than larger
ones; thus,

\vspace{-.1in}
\begin{eqnarray}
&\hskip-.4in\left[\matize A m\right]_{jk} &\hskip-.1in= a_{i_1\dots i_m\dots i_M},
\quad\mbox{where} \\
&&\hskip-.2in\quad j=i_m \quad \mbox{and} \quad k=1+\sum_{n=1\atop
n\neq m}^M(i_n - 1)\prod_{l=1\atop l\neq m}^{n-1} I_l.\nonumber
\label{eq:matrixizing}
\vspace{-.2in}
\end{eqnarray}
\label{def:matrixizing}
\end{definition}
%\fi
\vspace{-.15in}
\noindent
A generalization of the product of two matrices is the product of a
tensor and a matrix \cite{Delathauwer00a}.%Carroll80,\vspace{-.025in}
\vspace{-.025in}
\begin{definition}[Mode-$m$ Product, $\times\mode m$]
\label{def:mode-m-prod}
The mode-$m$ product of a tensor 
$\ten A \in \Reals^{I_1 \times I_2 \times \dots \times I_m \times \dots \times I_M}$ 
and a matrix 
$\mat B \in \Reals^{J_m \times I_m}$, 
denoted by $\ten A \times_m \mat B$, is
a tensor of dimensionality \\
$\Reals^{I_1 \times \dots \times I_{m-1}
\times J_m \times I_{m+1} \times \dots \times I_M}$ whose entries are
%computed by
\vspace{+.025in}
\begin{equation}
[\ten A \times_m \mat B]_{i_1 \dots i_{m-1} j_m i_{m+1} \dots i_M} =
\sum_{i_m} a_{i_1 \dots i_{m-1} i_m i_{m+1} \dots i_M} b_{j_m i_m}.
\nonumber\label{eq:mode-m-product}
\vspace{-.05in}
\end{equation}
\end{definition}
\noindent
The mode-$m$ product can be expressed in tensor notation, as
%\begin{equation}
$\ten C = \ten A \times_m \mat B,$
%\label{eq:mode-m-product-ten}
%\end{equation}
or in terms of 
%by 
 matrixized tensors, as
%\begin{equation}
$\matize C m = \mat B \matize A m.$
%\label{eq:mode-m-product-mat}
%\end{equation}
\iffalse
The mode-$m$ product can be expressed in matrix notation and tensor notation, as
\begin{equation}
\hskip-.525in \matize C m = \mat B \matize A m.
%\label{eq:mode-m-product-ten}
%\end{equation}
%or in terms of 
%by 
% matrixized tensors, as
%\begin{equation}
\hskip+.125in \Leftrightarrow\hskip+.2in 
\ten C = \ten A \times_m \mat B, 
\label{eq:mode-m-product-mat}
\end{equation}
\fi
%Briefly, an order $N>2$ tensor or $N$-way array
%$\mathcal D$ is an $N$-dimensional matrix comprising $N$ spaces.
The $M$-mode SVD (aka. the Tucker decomposition) %, Algorithm~\ref{alg:m-mode-svd},
is a ``generalization'' of the conventional matrix (i.e.,
2-mode) SVD which may be written in tensor notation as
\begin{eqnarray}
&\hskip-.3in \mat D&\hskip-.1in= \mat U_1 \mat S \mat U_2\tp \hskip+.225in \Leftrightarrow\hskip+.2in \mat D = \mat S \times_1 \mat U_1 \times_2 \mat U_2.
\end{eqnarray}
The $M$-mode SVD orthogonalizes the $M$ spaces and decomposes the
tensor as the {\it mode-m product}, denoted $\times_m$ 
%(see(\ref{eq:mode-n-product}))
, of $M$-orthonormal spaces, as follows:
\begin{eqnarray}
&\hskip-.3in \ten D&\hskip-.1in={\ten Z} \times_1 {\mat U}_1 \times_2 {\mat U}_2
\dots\times_m {\mat U}_m \dots \times_M {\mat U}_M.
\label{eq:tensor-decomposition}
\end{eqnarray}
\vspace{-.25in}
\vskip+0.4in
\begin{algorithm}[!b]
\begin{minipage}{\linewidth}
%% \hrule
%\medskip
%% \underline{\bf $M$-Mode SVD Algorithm:}
%\vspace{-.1in}
{\bf Input} the data tensor $\ten D\in\Reals^{I_1\times\dots\times
I_M}$.
\begin{enumerate}
\item
For $m \assign 1,\dots,M$,\\
Let $\mat U_m$ be the left orthonormal matrix of the SVD of
$\matize D m$, the mode-$m$ matrixized $\ten D$.
\item
Set $\ten Z \assign \ten D\times_1\mat U_1\tp \times_2 \mat U_2\tp
\dots\times_m \mat U_m\tp \dots \times_M \mat U_M\tp$.
%\medskip
\end{enumerate}

{\bf Output} mode matrices $\mat U_1,\dots,\mat U_M$ and the core
tensor $\ten Z$.\\
%\medskip
\end{minipage}
%\hrule
\caption[$M$-mode SVD algorithm]{$M$-mode SVD algorithm.}
\label{alg:m-mode-svd}
\vskip-.1in
\end{algorithm}

\vspace{-.2in}
\section{Global Tensor Factorization}
%\vskip-.05in
\label{sec:global-tensor}
\noindent
There are two classes of data tensor modeling techniques~\cite{Kolda09,Sidiropoulo2017} that stem from: the rank-$K$ decomposition (CANDECOMP/Parafac decomposition)~\cite{Carroll70,Harshman70,Bro97} and the multilinear rank-($R\mode 1,R\mode 2, \dots,R\mode M)$, such as Tucker decomposition~\cite{Tucker66}\cite{Delathauwer00a}\cite{desilva08}, such as Multilinear-PCA, multilinear (tensor) ICA, plus various kernel variations that are doubly nonlinear~\cite{Vasilescu09}.
%\clearpage
%\newpage
%\vskip-.1in

\subsection{\hspace{-0in}Representation: Multilinear Tensor Factorization}%  and Kernel Variants}
\vspace{-.05in}
Within the tensor mathematical framework, an ensemble of observations is organized in a higher order data tensor, $\ten D$.
Given a data tensor $\ten D$ of labeled, vectorized training facial images
$\vec d_{pvle}$, where the subscripts denote the causal factors of facial image
formation, the person $p$, view $v$, illumination $l$, and expression $e$ labels, the
$M$-mode SVD~\cite{Vasilescu03}\cite{Vasilescu02a}\cite{Delathauwer00a}\cite{DeLathauwer00b} %or the $M$-mode ICA algorithm~\cite{Vasilescu05} 
or its kernel variant~\cite{Vasilescu09} may be employed to multilinearly decompose the data tensor, 
%(Algorithm~\ref{alg:mpca}) 
\begin{equation}
     \ten D = \ten T \times\people \mat U\people \times\views \mat U\views \times\illums \mat U\illums \times\expres \mat U\expres,
\end{equation}
and compute the mode matrices $\mat U\people$, $\mat U\views$, $\mat U\illums$, and $\mat U\expres$ that span the causal factor representation.  The extended core computed  by, $\ten T=\ten D\times\people \mat
U\people\tp \times\views \mat U\views\tp \times\illums \mat
U\illums\tp \times\expres \mat U\expres\tp,$
%(\ref{eq:tensorfaces}) 
governs the interaction between the causal factors
(Figure~\ref{fig:tensorfaces}). This approach makes the assumption that the representations for a causal factor are well modeled by a Gaussian distribution.  An image $\vec d_{pvle}$ is represented by a person, view,
illumination and expression coefficient vectors as,
{%\small
\begin{equation}
\vec d_{pvle}=\ten T\times\people \vec p_p\tp \times\views \vec v_v\tp
\times\illums \vec l_l\tp \times\expres \vec e_e\tp.
\label{eq:mpca-rep}
\end{equation}}
An important advantage of 
employing vectorized observations in a multilinear tensor framework %ensemble of \\
%\underline{vectorized} observations 
is that all images of a person,
regardless of viewpoint, illumination and expression are mapped to the same person coefficient vector, thereby achieving zero intra-class scatter. Thus,
multilinear analysis creates well separated people classes by maximizing the ratio of
inter-class scatter to intra-class scatter.  
%This multilinear representation is illustrated in Figure~\ref{fig:tensorfaces}.
%(a for the data tensor in
%Figure~\ref{fig:datatensor}, which lacks the expression mode.
Alternatively, one can employ the Multilinear (Tensor) Independent Component Analysis (MICA) algorithm
%(Algorithm~\ref{alg:mica}), 
~\cite{Vasilescu05}, $M$-mode ICA (as opposed to the tensorized computation of the conventional, linear ICA\cite{Delathauwer97}), which takes advantage of higher-order
statistics~\cite{Common94} to compute the mode matrices that span the causal factor representation, and the MICA basis tensor that governs their interaction.
\noindent
\subsection{Recognition: Mutilinear Projection} 
%\vspace{-.05in}
While TensorFaces (MPCA)~\cite{vasilescu02,Vasilescu03} is a handy moniker for an approach that %employs second order statistics to 
learns from an image ensemble 
the interaction and representation of various causal factors that determine observed data, with Multilinear (Tensor) ICA~\cite{Vasilescu05} as a more sophisticated approach, %that employs higher order statistics,% to learn the interaction and representation of the causal factors of data formation.
none of the interaction models prescribe a solution for how one might determine the causal factors of a single unlabeled test image that is not part of the training set.  Multilinear projection~\cite{Vasilescu2011,Vasilescu07a} simultaneously projects one or more unlabeled test images that are not part of the training data set into multiple constituent causal factor spaces associated with data formation, in order to infer the mode labels:
%\vspace{-.15in}
\begin{equation}
\vspace{-.25in}
{\ten R} =\ten T\pinv{\pixels}\timesT\pixels {\vec d}\mode{new} \approx {\vec r}\people
\circ {\vec r}\views \circ {\vec r}\illums 
\circ {\bf r}\expres. 
\vspace{.275in}
\end{equation}
%where $\ten T\pinv{\pixels}$ is the pseudo-inverse in the pixel mode such that $\matize {T\pinv{}} \pixels$ 
The multilinear projection of a facial image computes the illumination, view and person representation by decomposing the expected rank-$1$ %(\ie rank-$(1,\dots,1)$) 
structure of the confounded response tensor ${\ten R}$, and taking advantage of the unit vector constraints associated with
${\vec r}\illums$, ${\vec r}\views$, ${\vec r}\expres$, and ${\vec r}\people$. 
%enables us to compute these three coefficient vectors 
via the CP-decomposition, \ie the rank-$1$ tensor decomposition.
%multilinear projection decomposes $\ten R$, a confounded representation of $\vec d\mode{new}$ into the individual causal factor representations.

\begin{figure*}%[!htbp]%\begin{figure*}[t]
%\begin{center}
\vspace{-.05in}
\centerline{
\begin{tabular}{l}
\includegraphics[width=.35\linewidth]{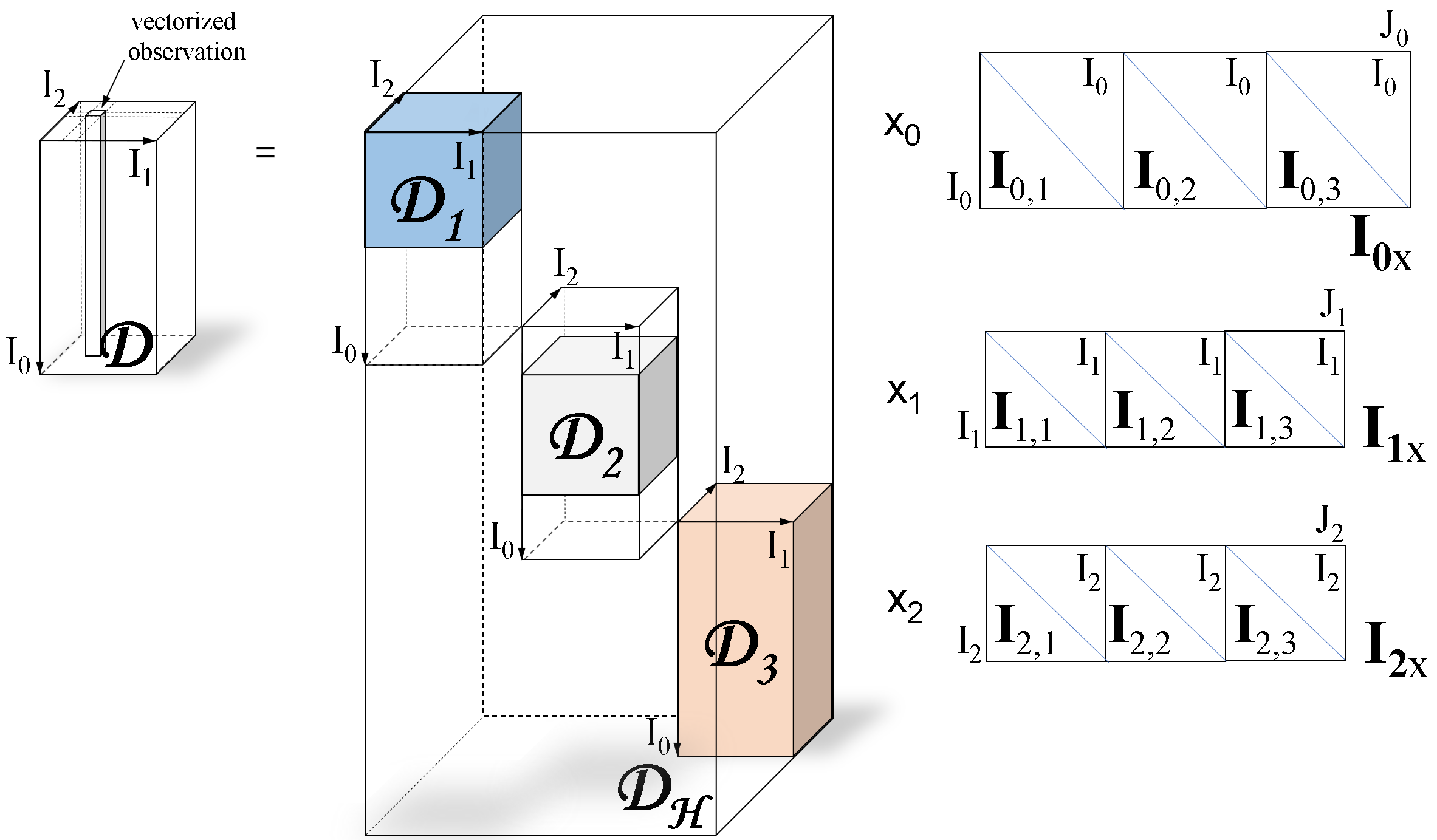}%{images/fully_compositional6.png}%{images/fully_compositional13.png}
\\
\end{tabular}
\hfill
%&
%\includegraphics[width=.45\linewidth]{images/Block-Tucker-Base-Case-general_parts_one_composite_mode_crop.png}}
%\hfill
\begin{tabular}{r}
\includegraphics[width=.47\linewidth]%{images/3layers_compositional2.png}
{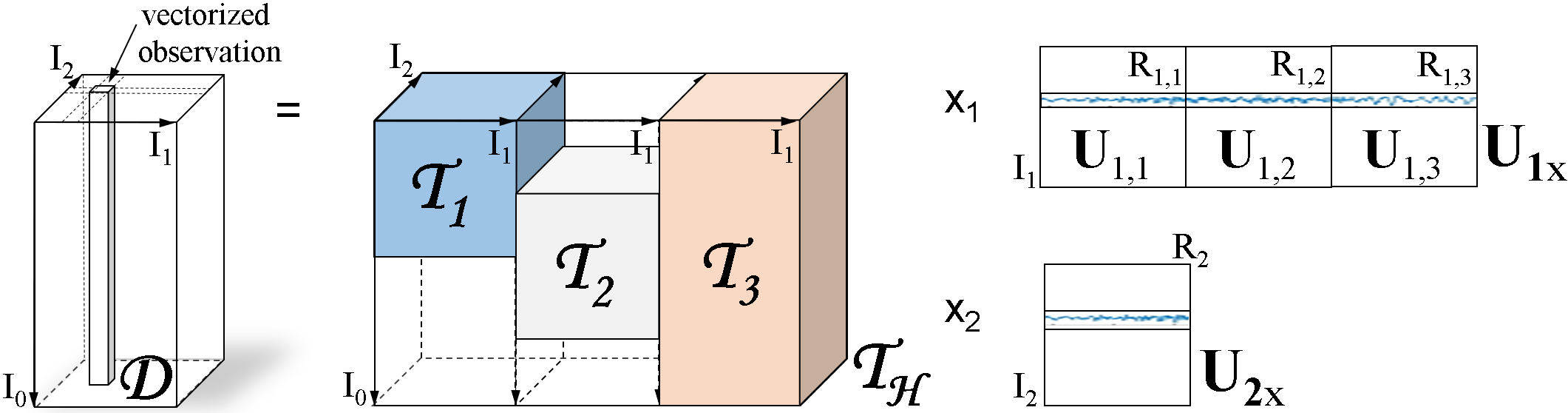}%{images/compositional_one_mode6.png}%{images/compositional_one_mode3.png}
%\hskip+.2in
\hfill
\\
\\
\end{tabular}}
\vspace{-.15in}
\centerline{\footnotesize \hspace{1.6in}(a)\hfill (b) \hspace{1.6in}}
%\centerline{\footnotesize \hspace{1.6in}\includegraphics[height=.013\linewidth]{images/arrow2.png}\hfill  \hspace{1.6in}}
%\centerline{\footnotesize \hspace{1.6in}V\hfill  \hspace{1.6in}}
%\vspace{-.02in}
%\centerline{ \hfill (d) \hfill (e)\hfill}
%\vspace{+.05in}
\centerline{
%\fbox{\rule{0pt}{2in} \rule{1.1\linewidth}{0pt}}
%\includegraphics[width=.8\linewidth]{images/Block_Tucker_General_Case_04.png}
%\includegraphics[width=1\linewidth,height=.165\linewidth]{images/Block-Tucker-Base-Case-general_parts_horizontal9_14.png}
\includegraphics[width=1\linewidth]
{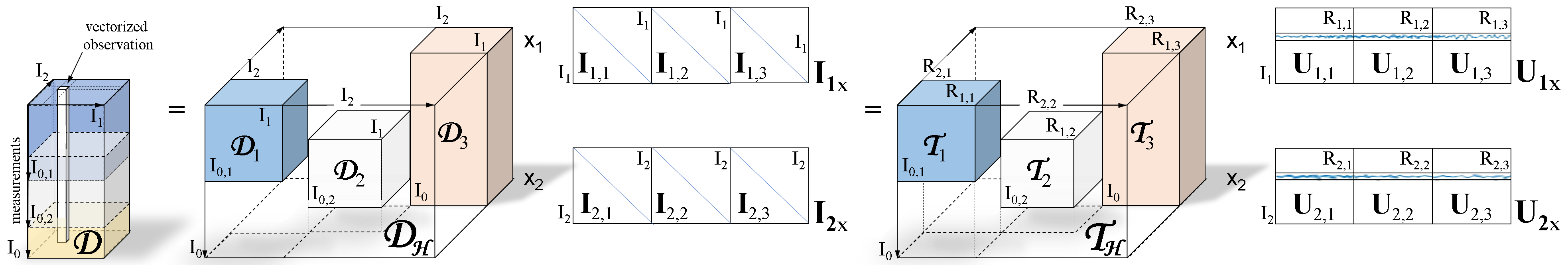}%{images/Block-Tucker-Base-Case-general_parts_horizontal9_11.png}}%{images/Block-Tucker-Base-Case-general_parts.png}}{images/Block-Tucker-Base-Case-general_parts.png}
}%\hfill
\vspace{-.075in}
\centerline{\footnotesize \hspace{1.6in}(c1) \hfill (c2)\hspace{1.65in}}
%\centerline{\footnotesize \hfill (e) \hfill}
%\vskip -.05in
%\vspace{-0.2in}
%\centerline{\footnotesize \hspace{1.6in}(d) \hfill (e)\hspace{1.65in}}

\vskip +.025in
%\end{center}
\centerline{
\hfill
\hskip -.325in
\includegraphics[height=.159\linewidth]{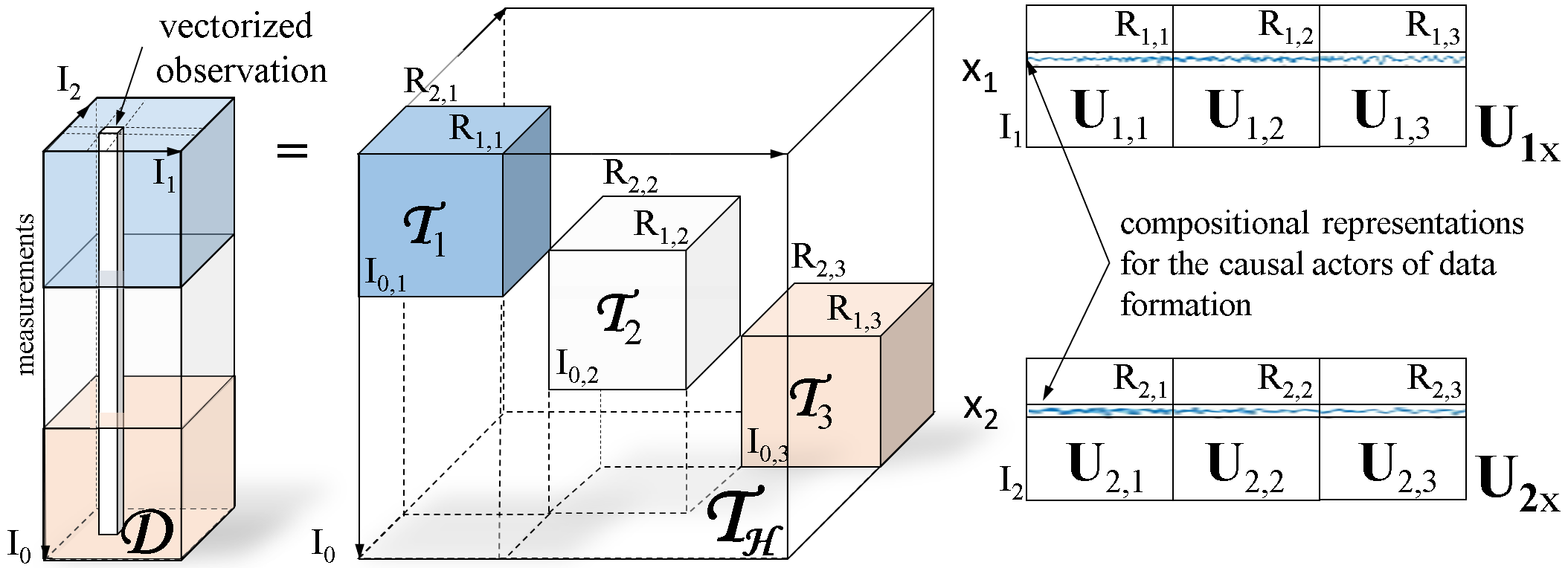}%{images/Block-Tucker-Base-Case-independent_parts7.png}%{images/Block-Tucker-Base-Case-independent_parts11.png}
\hfill
\hskip +.2in
\includegraphics[height=.165\linewidth]{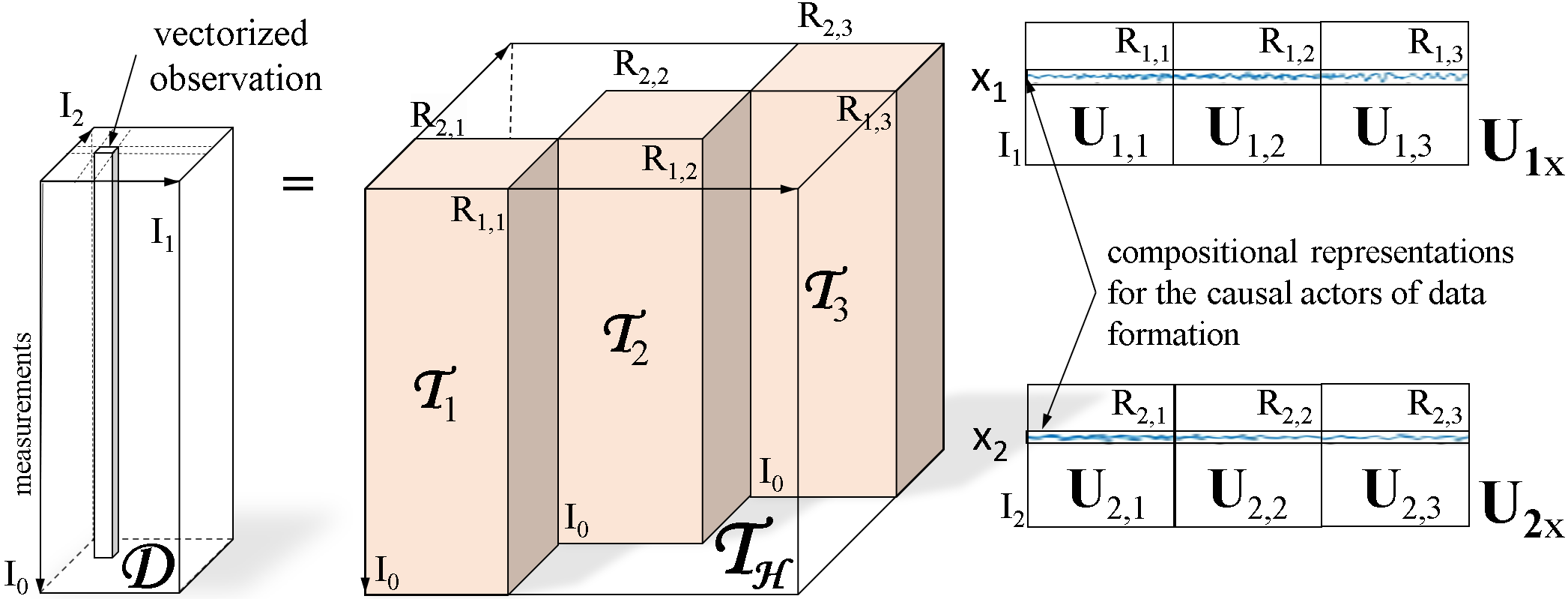}%{images/Block-Tucker-Base-Case-overlaping_parts7.png}%{images/Block-Tucker-Base-Case-overlaping_parts11.png}
\hfill
}
\vspace{-0.05in}
\centerline{\footnotesize \hspace{1.6in}(d) \hfill (e)\hspace{1.65in}}
\vspace{-.175in}
\caption[]{
Compositional Hierarchical Tensor Factorization.%($I\mode M$ refers to the measurement dimension, \ie number of pixels in an image. In the text, we denote the measurement dimension with $I\mode 0$.)%, a third order data tensor, $\ten D\in \Reals^{I\measure \times I_1 \times I_2}$, that contains a collection of vectorized observations $\vec d \in \Reals^{I\measure}$ that are the result of varying one causal factor $I\mode1$ times, while a second causal factor is varied $I\mode2$ times. For example, $\ten D$ could contains the vectorized facial images $I_1$ people photographed from $I_2$ view points and each image has a total of $I\measure$ pixels.
 %This a generalization of a Tucker-2 style block factorization. 
(a) The data tensor, $\ten D$, is rewriten in terms of a compositional hierarchical data tensor, $\ten D\tmode H$. A {\it fully compositional hierarchical data tensor} is a data tensor in which every mode is written in a compositional form. The compositional hierarchical data tensor, $\ten D\tmode H$, contains data tensor segments along its super-diagonal. %Notice, this figure depicts how the data tensor $\ten D$ has been rewritten; there is no factorization.  
%The data segments may be parts of $\ten D$ or a filtered version of the whole $\ten D$. 
 The data tensor segment, $\ten D_{s}$, may contain zeros and represent parts, or may be full and correspond to a filtered version of a parent-whole. 
%, $\ten D = \sum_{s=1}^S\ten D \times\mode M \mat H\mode s =\sum_{s=1}^S \ten D\mode s = \ten D{\tmode H} \times\mode M [\mat I\mode{M,1} \dots \mat I\mode {M,s} \dots \mat I\mode {M,S}] \times\mode 1 [\mat I\mode {1,1} \dots \mat I\mode {1,s} \dots \mat I\mode {1,S}] \dots \times\mode c [\mat I\mode {c,1} \dots \mat I\mode {c,s} \dots \mat I\mode {c,S}] \dots  \times\mode C [\mat I\mode {C,1} \dots \mat I\mode {C,s} \dots \mat I\mode {C,S}]$. 
(b) 
Only one of the causal factor has a compositional hierarchical representation, a {\it partially compositional hierarchical} representation example.% with o but not the rest.
%(c) 
%In the base case, the data tensor, $\ten D$, contains a collection of vectorized observations, where each observation is composed of a whole with two parts. The parts are overlapping and the whole may have measurements (\ie pixels) not contained in any of the parts. (
(c) In practice, the measurement mode will not be written in a compositional form, ie. $\ten D\tmode H$ would have already been multiplied with $\mat I\measure{}\mode{x}$%^{\ref{note:measurement_mode}}$%\footnote{$I\mode M$ refers to the measurement dimension, \ie number of pixels in an image. In the text, we denote the measurement dimension with $I\mode 0$.} 
 as in Fig.~\ref{fig:Block-Tucker-Base-Case}(c1). (c1) Rewriting the data tensor $\ten D$ as a hierarchical data tensor, $\ten D\tmode H$.
\iffalse
, and performing a single standard multilinear tensor decomposition will result in a part-based object representation. However, we show that there exist two more efficient ways to factorize $\ten D\tmode H$.  The compositional part-based representation may also be computed in an incremental-hierarchical manner. %The hierarchical data tensor, $\ten D{\tmode H}$, is a mathematical device that aids in deriving a compositional hierarchical factorization by placing the decomposition on solid mathematical foundation. 
%The measurements that correspond to a part may not be tightly packed into a block, but a permutation may be applied in order to chunk the data without changing the analysis. 
The parts $\ten D\mode s$ where $s=1,\dots, S$, have been extracted by applying a set of filters $\mat H\mode s$, %.which is on the left side if the filters are known, and on the right side if it ought to be estimated. 
followed by a permutation of the rows to chunk the data. This is a type-2 decomposition, \ie the measurement mode was not factorized.%The extended core ${\ten T}{\tmode H}$ governs the interaction between the causal factors that determine the observed data. 
\fi
(c2) Performing a compositional hierarchical tensor factorization results in a part-based causal factor representation, $\ten D=\ten T{\tmode H}\times\mode 1 \mat U\mode{1x} \times\mode 2 \mat U\mode{2x}$, where the extended core is $\ten T\tmode H=\ten Z\tmode H \times\measure \mat U\measure{}\pixels$.%, but computed according to eq.~\ref{eq:comp_tensor_model}. 
(d) Non-overlapping parts. The part representations' computations are independent of one another. (e) Completely overlapping parts.
%(f) A slightly more generic example is depicted in the appendix, Fig.~\ref{fig:3layers_generic}.
} 
%The parts may represent different frequencies, color channels, or various levels from a Laplacian pyramid, etc. 
%where each filter is implemented as doubly circulant matrix,  and the convolution operation is matrix-vector multiplication with a vectorized observation. In the base case, $\ten F$ contains three doubly circulant matrices. If the matrices $\mat H\mode{k}$ are have a limited scope and are block identity matrices, the filter matrix multiplication with a vectorized observation is equivalent to performing part segmentation. Moving the convolutional filter pyramid to the other side of the equation results in a dictionary of convolutional basis vectors.
\label{fig:Block-Tucker-Base-Case}
\label{fig:fully_compositional}
\label{fig:independentparts} 
\vspace{-.15in}
\end{figure*}

%\vskip -.15in
\section{Compositional Hierarchical Tensor Factorizations}
\label{sec:Compositional-Tensor-Factorization}
%\vskip -.05in
%While the MPCA or MICA compute object representations that are invariant of the extrinsic causal factors (illumination, imaging, \etc) that impact an object's appearance in an image, these methods compute global object features which are not robust to occlusion, and require a substantial amount of training data  in order to achieve the same performance as our proposed approach.
In prior tensor based research, an imaged object was represented in terms of global causal factor representations that are not robust to occlusions. 
This section introduces a compositional hierarchical tensor factorization %approach 
%and 
%explicitly represent an object's recursive hierarchy of wholes and parts.
%The compositional tensor factorization will be computed in a hierarchical manner where the whole representation computation is bootstrapped and computed based on the part representations. 
% of the wholes based on the intrinsic representation of the parts. % in terms of its hierarchy of intrinsic causal factors. %, and a dimensionality reduction algorithm that may be employed to suppress causal factors that hinder recognition.
%Therefore, this section  
%The compositional hierarchical factorization 
that derives its name from its ability to represent a hierarchy of intrinsic and extrinsic causal factors of data formation.%that determine image appearance.% which will address in this paper.  %A hierarchical computation of the factorization will be addressed in a seperate paper.
% Representing a ierarchy of intrinsic and extrinsic causal factors of object appearance 
A hierarchical representation may be effectively employed to recognize occluded objects, including self-occlusion that occurs during out-of-plane rotation relative to the camera viewpoint. The efficacy of our approach is demonstrated by our LFW and Freiburg verification experiments that compare global and hierarchical representations in Section~\ref{sec:Compositional-TensorFaces}. 
     
\iffalse
\begin{enumerate}
    \item 
    %the representation of an object based on a hierarchy of intrinsic and extrinsic appearance based properties, ie 
     hierarchy of intrinsic and extrinsic causal factors that determine the appearance of an object that may be employed more effectively to recognize objects when occlusion, including self-occlusion that takes place during out-of-plane rotation as camera changes view point.
    \item
    a hierarchical computation of the representation where the parent-whole representation is bootstrapped and computed based on the pre-computed child-part representations. 
\end{enumerate}
\fi
%This paper addresses the first 
%In prior research, an object was represented by computing global %intrinsic representations that were invariant of extrinsic factors . In %this section, we introduce a compositional tensor factorization paradigm for explicitly representing an object’s recursive hierarchy of wholes and parts, and a156hierarchical computation of the intrinsic and extrinsic causal factors of the wholes based on the intrinsic representation of the parts

%\subsection{Hierarchical Data Tensor}
%\subsection{Mathematical Basis:}
%{Representing an Object's Recursive Hierarchy of Wholes and Parts:}

Within the tensor mathematical framework, an ensemble of training observations is organized in a higher order data tensor, $\ten D$.
%We will consider the decomposition of a higher order training data tensor, %Fig.~\ref{fig:Block-Tucker-Base-Case}, 
A data tensor $\ten D\in\Reals^{I\measure \times I\mode1 \times I\mode c \dots \times I\mode C}$ 
%where there are $I_0$ measurements, the result of $C$ causal factors and the $c\th$ causal factor was varied $I_c$ times. 
% that 
contains a collection of 
 %where a 
 vectorized observation, $\vec d_{i\mode1\dots i\mode c\dots i\mode C} \in\Reals^{I\measure}$ where each subscript, $i_c$, %with values $1\le i_c \le I_c$
 denotes one of the $C$ causal factors that have created the observation and have resulted in $I_0$ measurements, \ie a total of $I\measure$ pixels per image. 
 %, where $1\le i\mode c \le I_c$, and have resulted in $I\measure$ measurements
In this paper, we will report results based on a data tensor $\ten D$ of labeled, vectorized training facial images
$\vec d_{pvle}$, where the subscripts denote the causal factors of facial image
formation, the person $p$, view $v$, illumination $l$, and expression $e$ labels.

\vspace{-.05in}
\subsection{Hierarchical Data Tensor}
\vspace{-.05in}
%\section{Compositional Hierarchical Factorization}
We identify a general base case object and two special cases whose intrinsic and extrinsic causal factors we would like to represent. An object may be composed of
%An observation could be made up of either 
(i) two partially overlapping children-parts and parent-whole that has data not contained in any of the children-parts, 
(ii) a set of non-overlapping parts, or
(iii) a set of fully overlapping parts,  which resembles to the rank-$(L,M,N)$ or a rank-$(L,M,\cdot)$ block tensor decomposition~\cite{DeLathauwer08}, but which is too restrictive for our purpose.\footnote{
%The block tensor decomposition is too restrictive for our purpose. %and it was defined as finding the minimum number of fully overlapping 
The block tensor decomposition~\cite{DeLathauwer08} computes the best fitting $R$ %rank-(R_1, R_2, R_3) 
fully overlaping tensor blocks that are all multilinearly decomposable into the same multilinear rank-(L,M,N),%, ie the minimum number of $R$ fully overlapping multilinearly block decomposable parts with the same that 
which is analogous to finding the best fitting $R$ rank-$1$ terms computed by the CP-algorithm.
}
% However, the algorithm described~\cite{DeLathauwer08c} computed the best fitting $K$ multilinear blocks for a user specified $K$.}
 Figure~\ref{fig:Block-Tucker-Base-Case} depicts the general base case and the two special base cases.
%In the general case, an observation, $\vec d$, is a recursive hierarchy of wholes and parts with partly overlapping children parts, and 
%a parent whole that may have measurements not contained in any of its parts.  
In real scenarios, parent-wholes have children-parts that are recursively composed of children themselves, Fig.~\ref{fig:HTensorFaces}.

The data wholes, and parts are extracted by employing a filter bank, $\{\mat H\mode s \| \mat H\mode s\in \Complex^{J\measure\times I\measure}, 1 \le s \le S\}$ where
%\footnote{The filters can be quickly adjusted by pre-multiplying them with weighting matrix, $\mat W$, where $\mat W \assign (\sum_{s=1}^S \mat H\mode s)^{-1}$ in order to satisfy $\sum_{s=1}^S \mat H\mode s = \mat I$ constraint.}.
 each $2D$ convolutional filter is implemented as a doubly (triply) block circulant matrix,$\mat H\mode s$, %\footnote{A $1D$ and a $3D$ filter are implemented as a circulant matrix and triply block circulant matrix, respectively.}
 where 
 $\sum_{s=1}^S \mat H\mode s = \mat I$, and $s$ refers to the data segment. Convolution is a matrix-vector multiplication,
%\begin{eqnarray}
%\ * \
%\end{eqnarray}
between a circulant matrix and a vectorized observation, $\vec d\mode s=\mat H\mode s\vec d$ which in tensor notation is written as $\vec d\mode s=\vec d\times\measure \mat H\mode s$ where mode $0$ is the measurement mode.
%, that in practice is efficiently implemented using a FFT. 
%An object segment (whole, part or subpart), $s$, is extracted by ''convolving'' the entire observation with the filter. 
% A filter is implemented as a doubly block circulant matrix,$\mat H\mode s$, for a $2D$ convolution filter\footnote{A $1D$ and a $3D$ filter are implemented as a circulant matrix and triply block circulant matrix, respectively.}, and 
% a convolution operation is a matrix-vector multiplication between a matrix filter and a vectorized observation, $\vec d$. 
%In the our base case general example, $\ten F$ contains three doubly block circulant matrices.  
The segment data tensor, $\ten D_s=\ten D \times\measure \mat H\mode s$, is the result of multiplying (convolving) every observation, $\vec d$, with the block circulant matrix (filter), $\mat H\mode s$.  A  filter  $\mat H\mode s$  may be any of any type, or have any spatial scope. %, where ${\mat H\mode g | 1\le g\le G}$.%, and a filter tensor bank $\ten F$ may implement any type of filter.
When a filter matrix 
is a block identity matrix, $\mat H\mode{s}=\mat I\mode s$, the filter matrix multiplication with a vectorized observation has the effect of segmenting a portion of the data without any blurring, subsampling or upsampling.%\footnote{Subsampling may be implemented by dropping every other row from the filter matrix.} 
Measurements associated with perceptual parts may may not be tightly packed into a block a priori, as in the case of vectorized images, but chunking may be achieved by a trivial permutation.

The data tensor, $\ten D$, is expressed
in terms of its recursive hierarchy of wholes and parts
by defining and employing a {\it hierarchical data tensor}, $\ten D\tmode H$, %a mathematical representation of a tree data-structure 
that contains along its super-diagonal the collection of wholes and parts, $\ten D\mode s$, Fig.~\ref{fig:fully_compositional}(a)
%\footnote{\label{note:measurement_mode}. In the figures, $I\mode M$ refers to the measurement dimension, \ie number of pixels in an image. In the text, we denote the measurement dimension with $I\mode 0$.}, 
%, ${\ten D}{\tmode H}$, with the expression%data tensor, $\ten D{\tmode H}$ as %as unified model of wholes and parts, 
%as a hierarchical data tensor model
%where $\{\ten D\mode s \|\ten D\mode s = \ten D \times\mode M  \mat H\mode s, \mbox{ where } 1\le s %\le S\}$.
\\

\vspace{-.2in}
\begin{eqnarray}
\ten D &=&
\sum_{s=1}^S%{\hskip+.22in S}\mathop{}_{\hskip-.075in s=1}
\ten D \times\mode 0 \mat H\mode s \label{eq:D_conv}\\
&=&
%\underbrace{
\ten D\mode {1} \dots + \ten D\mode {s}  \dots + \ten D\mode {S}%}_{(b)}
\label{eq:D_segment_rep}\\
%\nonumber \\
%\vspace{-.5in}
%^\begin{empheq}
%[box=\fbox]{align}
%\ten D 
\hskip -.1in &=& %\hskip -.1in
%&=&
%\underbrace{
\ten D{\tmode H} \times\mode 0 \mat I\mode{Ox}\times\mode 1 \mat I\mode{1x} \dots \times\mode c \mat I\mode{cx} \dots \times\mode C \mat I\mode{Cx}%}_{(c)}
,\label{eq:DF_rep}
\vspace{-.35in}
%\end{empheq}
\end{eqnarray}

\noindent
%where $\ten D{\tmode H}$ is a {\it hierarchical data tensor} that contains an object's recursive hierarchy, the collection of wholes and parts, $\ten D\mode s=\ten D \times\measure \mat H\mode s$, along its super-diagonal, Fig.~\ref{fig:fully_compositional}(a), 
where $\mat I\mode{cx}=[ \mat I\mode {c,1} ... \mat I\mode {c,s} ... \mat I\mode {c,S}]\in\Reals^{I_c\times SI_c}$ is a concatenation of $S$ identity matrices, one for each data segment. %A hierarchical data tensor is a mathematical representation of a tree data-structure. 
The three different ways of rewritting $\ten D$ in terms of a hierarchy of wholes and parts, eq.~\ref{eq:D_conv}-\ref{eq:DF_rep}, results in three equivalent
 compositional hierarchical tensor factorizations~\footnote{\label{note:equivalence}Equivalent representations can be transformed into one another by post-multiplying mode matrices with permutations or more generally nonsingular matrices, $\mat G\mode c$,\\ $\ten D= ({\ten Z}_{\ten H}\times\mode 0 \mat G\inv\mode 0 \dots \times\mode c \mat G\inv\mode c \dots \times\mode C \mat G\inv\mode C) \times\mode 1  \mat I\mode{1x}{\mat U\mode 1}{}_{\ten H}\mat G\mode 1 \dots \times\mode c  \mat I\mode{cx}{\mat U\mode c}_{\ten H} \mat G\mode c \dots \times\mode C  \mat I\mode{Cx}{\mat U\mode C}_{\ten H} \mat G\mode C$.}  %These three different are equivalent up to a permutation or a nonsingular transformation, The latter one decomposition is computationally and memory efficient.
%, and  
%The data tensor, $\ten D$, is not re-organized into $\ten D\tmode H$. 
%where a recursive hierarchy of wholes and parts 
%of a collection of observations organized in a data tensor, $\ten D$ 
%The hierarchy is explicitly represented in terms of their statistically invariant intrinsic properties.
\vspace{-.1in}
\begin{eqnarray}
\hskip-.8in\ten D 
&\hskip-.1in=&\hskip-.1in
\sum_{s=1}^S%{\hskip+.22in S}\mathop{}_{\hskip-.075in s=1}
\underbrace{(\ten Z\times\measure \mat U\measure \times \mode 1 \mat U\mode 1 \dots \times\mode c \mat U\mode c \dots \times\mode C \mat U\mode C )}_{\mbox{%\scriptsize M-mode SVD of 
$\ten D$}}\times\mode 0  \mat H\mode s \label{eq:D_decomp}\\
&\hskip-.1in=&\hskip-.1in
\sum_{s=1}^S%{\hskip+.22in S}\mathop{}_{\hskip-.075in s=1}
\underbrace{(\ten Z\mode s\times\measure \mat U\measure{}\mode{,s} \times \mode 1 \mat U\mode {1,s} \dots \times\mode c \mat U\mode {c,s} \dots \times\mode C \mat U\mode {C,s} )}_{\mbox{%\scriptsize M-mode SVD of 
$\ten D\mode s$}}
\label{eq:Ds_decomp}\\
&\hskip-.1in=&\hskip-.1in
%&=&
%\sum_{s=1}^S%{\hskip+.22in S}\mathop{}_{\hskip-.075in s=1}
\underbrace{(\ten Z\tmode H\times\measure \mat U\measure{}\tmode H \times \mode 1 \mat U\mode {1}{}\tmode H \dots \times\mode c \mat U\mode {c}{}\tmode H \dots \times\mode C \mat U\mode {C}{}\tmode H)}_{\mbox{%\scriptsize M-mode SVD of 
$\ten D\tmode H$}}\nonumber\\
&&\hskip+.2in\times\mode 0 \mat I\mode{0x}\times\mode 1 \mat I\mode{1x} \dots \times\mode c \mat I\mode{cx} \dots \times\mode C \mat I\mode{Cx}%}_{(c)}
\label{eq:DH_decomp}
%\vspace{-.5in}
%\end{empheq}
\end{eqnarray}
\iffalse
These equation represent the $M$-mode SVD of $\ten D$, $\ten D\mode s$ and $\ten D\tmode H$ respectively. The summations in eq.~\ref{eq:D_decomp} and ~\ref{eq:Ds_decomp} can be rewritten in the hierarchical form similar to eq.~\ref{eq:DH_decomp} by setting $\mat U\mode{cx}$ to the matrix that contains the matrix $\mat U\mode c$ of eq.~\ref{eq:D_decomp} concatenated $S$ times and by setting $\ten Z\tmode H$ to the tensor that contains $\ten Z$ concatenated $S$ times along the super-diagonal. 
\fi
The expression of $\ten D$ in terms of a hierarchical data tensor is a mathematical conceptual device,
%\footnote{It is a mathematical representation of a tree data-structure, %~\footnote{It is not the intention to organize the data tensor, $\ten D$, into a hierarchical data tensor, $\ten D\tmode H$.} 
 that enables a unified mathematical model of wholes and parts
that can be expressed completely as a mode-m product (matrix-vector multiplication) and whose factorization can be optimized in a principled manner. 
% instead of employing a myriad of engineering solutions. 
Dimensionality reduction of the compositional representation is performed by optimizing 
%the approximation of $\ten D$ %and not $\ten D\tmode H$\footnote{Note, our goal is not to find the optimal multilinear representation of $\ten D\tmode H$ representation, but the optimal compositional multilinear representation of $\ten D$.  There is subtle but important difference.} 

%~\footnote{We will rewrite and amend this loss function in eq.~\ref{eq:loss_fnc} and eq.~\ref{eq:loss_fnc_rec} for computational purposes and to meet recognition goals.}
\vspace{-.1in}
\begin{eqnarray}
&\hskip-.4in e& 
%    \hskip -.1in &=& \hskip -.1in
%    \frac{1}{2} \|\ten D - \tilde{\ten D}\tmode H \times\mode 1 \mat I\mode{1}{}\pixels \dots
%    \times\mode 1 \mat I\mode{c}{}\pixels \dots
%    \times\mode 1 \mat I\mode{C}{}\pixels
%    \|^2 %\nonumber \\
%        %\hskip -.1in & & \hskip -.1in 
%5    + \sum_{c=1}^C \lambda\mode {c}\|\tilde{\mat U}\mode {c}{}\tmode H\tp {\tilde{\mat U}\mode c}{}\tmode H-\mat I\| 
    %+ \gamma\|\sum_{s=1}^S \mat H\mode s - \mat I\|
%    +\sum_{c=1}^C\sum_{s=1}^S\lambda\mode {c,s}\|\tilde{\mat U}\mode {c,s}\tp\tilde{\mat U}\mode {c,s}-\mat I\|
%\nonumber \\
    \hskip -.25in = \hskip -.125in
    %\hskip-.45in
    =
    \frac{1}{2} \|\ten D \hskip-.025in- \hskip-.025in
    (
\hskip+.025in \bar{\ten Z}\tmode H  \times\measure \bar{\mat U}\mode 0{}\tmode H %\times\mode 1 \tilde{\mat U\mode {1}}{}\tmode {H} 
... \times\mode c \bar{\mat U}\mode {c}{}\tmode H ...  \times\mode C \bar{\mat U}\mode{C}{}\tmode H)\nonumber \\
%\hskip -.15in & & \hskip -.15in 
& &\hskip+.2in 
\times\mode 0 \mat I\mode{0x} %\times\mode 1 \mat I\mode{1x} 
... \times\mode c \mat I\mode {cx}
... \times\mode C \mat I\mode {Cx}%\hspace{+.1in}
\|^2
\label{eq:loss_fnc}%\\
%\nonumber \\
%    \hskip -.1in & & \hskip -.1in 
    + \sum_{c=0}^C \lambda\mode {c}\|{{\bar{\mat U}\mode c}{}\tp\tmode H}{\bar{\mat U}\mode c}{}\tmode H -\mat I\|^2 %\nonumber
\label{eq:loss_fnc_DH}
\vspace{-.5in}
\end{eqnarray}
\noindent
\iffalse
\\
&\hskip -.2in =& \hskip -.125in
    \frac{1}{2}{} \|\ten D\tmode{H} - 
    (
\tilde{\ten Z}{\tmode H}  
%%%\times\measure \mat I\measure\pixels \mat H\tmode{H}\tilde{\mat U}\measure\tmode{H} 
%%%    \tilde{\ten T}{\tmode H}  
    \times\mode 0 {\tilde{\mat U}\mode {0}}{}\pixels ...%\dots 
    \times\mode c {\tilde{\mat U}\mode {c}}{}\pixels ...%\dots  
    \times\mode C {\tilde{\mat U}\mode{C}}{}\pixels)
%%%    \times\mode1 \mat I\mode{1x}...
%%%    \times\modec \mat I\mode{cx}...
%%%    \times\modeC \mat I\mode{Cx}
\|}^2
\nonumber \\
\hskip -.1in & & \hskip -.1in 
   \hskip -.05in + \hskip -.05in 
   \sum_{s=1}^S \sum_{c=1}^C \lambda\mode {c}{\|{{\tilde{\mat U}\mode {c,s}}{}\tp}{\tilde{\mat U}\mode {c,s}}-\mat I\|}^2 %\nonumber
%%    %+ \gamma\|\sum_{s=1}^S \mat H\mode s - \mat I\|
\fi
\vspace{-.1in}

\noindent
where $\bar{\mat U}\mode{c}{}\tmode{H}$ is the
composite representation of the $c^{\mbox{\small th}}$ mode, and $\ten Z\tmode H$ governs the interaction between causal factors.%, for $0\le c\le C$. 
\footnote{In the face recogniton application we discuss later, $c=0$ refers to the measurement mode, \ie the pixel values in an image, and the range of values $1\le c\le C$ refers to the $C$ causal factors.} Our optimization may be \underline{initialized},
 only, by setting $\ten Z\tmode H$ and $\mat U\mode c{}\tmode H$ to the M-mode SVD of $\ten D\tmode H$,
 \footnote{For computational efficiency, we may perform M-mode SVD on each data tensor segment  $\ten D\mode s$ and concatenate terms along the diagonal of $\ten Z\tmode H$ and $\mat U\mode c{}\tmode H$. However, the most computational efficient initialization, first, computes the M-mode SVD of $\ten D$, multiplies ($\ie$,convolves)
 the core tensor, $\ten Z$ with $\mat H\mode s$, followed by a concatenation of terms
 %and then concatenates terms 
 along the diagonal of $\ten Z\tmode H$ and duplication of $\mat U\mode c$ along the diagonal of $\mat U\mode c{}\tmode H$. The last initialization approach
 makes segment specific dimensionality reduction problematic, since part-based standard deviation, 
 %$\sigma{{}_{i}}_{c}{{}_{i}}_{c}{{}_{,s}}$, 
$\sigma_{i_ci_c,s}$, 
 is not computed.} 
 \iffalse
 \footnote{In the face recogniton application we discuss later, $c=0$ refers to the measurement mode, \ie the pixel values in an image, and the range of values $1\le c\le C$ refers to the $C$ causal factors.} Our optimization may be \underline{initialized},
 only, by setting $\ten Z\tmode H$ and $\mat U\mode c{}\tmode H$ to the M-mode SVD of $\ten D\tmode H$,
 \footnote{For computational efficiency, we may perform M-mode SVD on each data tensor segment  $\ten D\mode s$ and concatenate terms along the diagonal of $\ten Z\tmode H$ and $\mat U\mode c{}\tmode H$. This concatenation of terms corresponds to writing eq.~\ref{eq:Ds_decomp} in a hierarchical form and can be transformed${}^{\ref{note:equivalence}}$ into eq.~\ref{eq:DH_decomp} by employing a set of permutation matrices $\mat G\mode c$. However, the most computational efficient initialization, first, computes the M-mode SVD of $\ten D$, multiplies ($\ie$,convolves)
 the core tensor, $\ten Z$ with $\mat H\mode s$, followed by a concatenation of terms
 %and then concatenates terms 
 along the diagonal of $\ten Z\tmode H$ and duplication of $\mat U\mode c$ along the diagonal of $\mat U\mode c{}\tmode H$. This corresponds to writing eq.~\ref{eq:D_decomp} as a hierarchical form. The last initialization approach
 makes segment specific dimensionality reduction problematic, since part-based variance, 
 %$\sigma{{}_{i}}_{c}{{}_{i}}_{c}{{}_{,s}}$, 
$\sigma_{i_ci_c,s}$, 
 is not computed.} 
 \fi
 and performing dimensionality reduction through truncation, where $\bar{\mat U}\mode c{}\tmode H \in 
 \Reals^{SI_c \times \bar{J}_c}$, 
 $\bar{\ten Z}\tmode H \in \Reals^{\bar J_ 0 \dots \times \bar J_c \dots \times \bar J_C}$ and $\bar J\mode c \le SI\mode c$.
%Although $\ten D= \ten D\tmode H \times\mode 0 \mat I\mode{0x} \times\mode 1 \mat I\mode{1x} \dots \times\mode c \mat I\mode{cx} \dots\times\mode C \mat I\mode{Cx}$, moving  known quantities $\mat I\mode{cx}$ to the other side of equation is only approximately equal to $\ten D\tmode H$ ($ \ten D\tmode H \approx \ten D\times\mode 0 \mat I\mode{0x}\pinv{} \dots \times\mode{c} \mat I\mode{cx}\pinv{} \dots \times\mode{C} \mat I\mode{Cx}\pinv{}$) unless the hierarchy is made up of one data segment, \ie $\ten D=\ten D\tmode H$.
\iffalse
\begin{eqnarray}
\ten D \hskip-.05in&=& \hskip-.05in\ten D\tmode H \times\mode 0 \mat I\mode{0x}\dots \times\mode c \mat I\mode{cx} \dots \times\mode C \mat I\mode{Cx}\\
\ten D\times\mode 0 \mat I\mode{0x}\pinv{} \dots \times\mode{c} \mat I\mode{cx}\pinv{} \dots \times\mode{C} \mat I\mode{Cx}\pinv{}\hskip-.05in&\ne&\hskip-.05in \ten D\tmode H
\end{eqnarray}
\fi
%Thus, optimizing equation (\ref{eq:loss_fnc_DH}) directly will result in superior results.

\iffalse
In the rest of the paper, we will assume that only the causal factors modes, $c$ for $1\le c \le C$, have been written in a compositional form, \ie ${\ten D}\tmode H$ has already been  multiplied with ${\mat I}\measure{}\mode{x}$, and has a form as seen in Fig.~\ref{fig:Block-Tucker-Base-Case}(c1). 
%For clarity, we will identify the measurement mode with the letter $M$ and the mode matrix that spans the measurement space, ${{\mat U}\mode 0}{}\tmode H$, 
%will be denoted with ${{\mat U}\measure}{}\tmode H$. 
\fi
\noindent
\noindent
\subsection{Compositional Hierarchical Factorization Derivation}
%\vspace{-.05in}
%***Currently editing this section.*** 
%In order to take advantage of the data tensor segment structure, 
For notational simplicity, 
we %multiplying through the matrices $\mat I\mode{cx}$, and 
re-write the loss function as,%, eq.~\ref{eq:loss_fnc_DH},  optimize it by performing alternating least squares, minimizing for each parameter while holding the rest fixed,
%\vspace{-.15in}
\begin{eqnarray}
    e %\hskip -.1in &\assign& \hskip -.1in
    %\frac{1}{2} \|\ten D - \tilde{\ten D}\|^2 +\sum_{c=1}^C\sum_{s=1}^S\lambda\mode {c,s}\|\tilde{\mat U}\mode {c,s}\tp\tilde{\mat U}\mode {c,s}-\mat I\|
    \hskip -.1in &\assign& \hskip -.1in
    \frac{1}{2} \|\ten D - 
    \tilde{\ten Z}{\tmode H} \times\mode 0 \tilde{\mat U}\mode {0x}
...%
%\dots 
\times\mode c \tilde{\mat U}\mode {cx}
...%\dots 
\times\mode C \tilde{\mat U}\mode {Cx}
%   \sum_s(
%    \tilde{\ten Z}\mode s%\times\measure \mat H\mode {s}
%    \times\mode {0} \tilde{\mat U}\mode {0,s}
%    \times\mode {1} \tilde{\mat U}\mode {1,s}
%    ...%\dots 
%    \times\mode {c} \tilde{\mat U}\mode {c,s} 
%    ...%\dots 
%    \times\mode {C} \tilde{\mat U}\mode {C,s})
    \|^2 \nonumber\\
    \hskip -.1in & & \hskip -.1in 
    + \sum_{c=0}^C\sum_{s=1}^S \lambda\mode {c,s}\|\tilde{\mat U}\mode {c,s}\tp \tilde{\mat U}\mode {c,s}-\mat I\| %+ \gamma\|\sum_{s=1}^S \mat H\mode s - \mat I\|
    \label{eq:loss_fnc}
    \vspace{-.5in}
\end{eqnarray}

%\\
\noindent
%\vspace{-.2in}
%where ${\mat U}\mode {cx} = \mat I\mode{cx}{\mat U}\mode c{}\tmode H=[{\mat U}\mode{c,1}| \dots |{\mat U}\mode{c,s}| \dots |{\mat U}\mode{c,S}]$. % and ${\mat G}\mode c\in\Reals^{\bar J_c \times SI_c}$ is permutation matrix that groups the columns of $\mat U\mode c{}\tmode H$ based on the segment, $s$, to which they belong, and the inverse have been multiplied into $\ten Z\tmode H$ resulting into a $\ten Z\tmode H$ that has a set of basis vectors that have also been grouped based on their segments.  
where $\tilde{\mat U}\mode {cx} = \mat I\mode{cx}\bar{\mat U}\mode c{}\tmode H \tilde{\mat G}\mode c=[\tilde{\mat U}\mode{c,1}| \dots |\tilde{\mat U}\mode{c,s}| \dots |\tilde{\mat U}\mode{c,S}]$ and $\tilde{\mat G}\mode c\in\Reals^{\bar J_c \times SI_c}$ is permutation matrix that groups the columns of $\mat U\mode c{}\tmode H$ based on the segment, $s$, to which they belong, and the inverse permutation matrices have been multiplied${}^{\ref{note:equivalence}}$ into $\tilde{\ten Z}\tmode H$ %=\bar{\ten Z}_{\ten H}\times\mode 0 \tilde{\mat G}\inv\mode 0 ... \times\mode c \tilde{\mat G}\inv\mode c ... \times\mode C \tilde{\mat G}\inv\mode C$ 
resulting into 
%a $\ten Z\tmode H$ that has 
%a set of basis vectors that have
a core that has  also been grouped based on %their 
segments and sorted based on variance.  
\noindent
%\begin{figure}
%\centerline{
%  %\fbox{\rule{0pt}{2in} \rule{0.9\linewidth}{0pt}}
%  %\hfill
%  \includegraphics[width=1\linewidth]{images/Block-Tucker-Base-Case-overlapping_parts.png}
%  %\hfill
%  %\includegraphics[height=0.5\linewidth]{images/pope.png}
%  %\includegraphics[height=0.5\linewidth]{images/dog_parts_border.png}
%  %\includegraphics[height=0.475\linewidth]{images/dog.png}
%%\includegraphics[height=0.5\linewidth]{images/dog.png}
%  %\includegraphics[width=0.6\linewidth]{egfigure.eps}
%}
%\caption{Combinatorial tensor factor0002ization with overlaping parts .}  %\label{fig:independentparts} 
%\end{figure}
%\noindent
The data tensor, $\ten D$, may be expressed in matrix form as in eq.~\ref{eq:Dapprox_matrix1} and reduces to 
the more efficiently block structure as in eq.~\ref{eq:Dapprox_matrix2}
\vspace{-.06in}
%\rule{\dimexpr(.5\textwidth-1\columnsep-0.4pt)}{0.4pt}%
%\rule{0.4pt}{6pt}
%\vspace{-.2in}
%\begin{strip}
%\begin{align*}
\begin{eqnarray}
&\hskip-.25in {\ten D}
&\hskip -.15in = %\hskip -.15in
{\ten Z}\tmode H\times\mode 0 {\mat U}\mode{0x} \times\mode 1 {\mat U}\mode{1x}\dots \times\mode c {\mat U}\mode{cx}\dots \times\mode C {\mat U}\mode{Cx}
\\
\nonumber\\
&\hskip-.15in 
{\mat D}\mode{[c]} &\hskip-.15in 
={\mat U}\mode {cx} {\mat Z}\tmode{H}{}\mode{[c]}
\left({\mat U}\mode {Cx} \otimes \dots \otimes {\mat U}\mode{(c+1)x} \otimes {\mat U}\mode{(c-1)x} \otimes \dots \otimes {\mat U}\mode{0x}\right)\tp \hskip-.3in\label{eq:Dapprox_matrix1}\\
&\hskip-.15in
=& \hskip-.15in
%&\hskip-.1in =&
%=%\hskip -.15in 
\left[\hskip+.025in {\mat U}\mode{c,1} \dots {\mat U} \mode{c,s} \dots {\mat U} \mode{c,S}\hskip+.025in\right]\hskip-.3in\label{eq:Dapprox_matrix2}%\\
%&\hskip -.3in&\hskip -.3in 
\\
&&\hskip-.15in
\left[\hskip -.055in
\begin{array}{ccccc}
{\mat Z}\mode{0[c]} &\hskip-.1in\mat 0& \hskip -.1in&  \hskip -.1in\cdots&\hskip-.1in\mat 0 \\
\mat 0 &\hskip-.1in\ddots&  \hskip -.1in\mat 0&  \hskip -.1in &\hskip-.1in\vdots  \\
\vdots  &\hskip-.1in\mat 0 & \hskip -.1in{\mat Z}\mode{s[c]}& \hskip -.1in \mat 0 & \hskip-.1in \\
 &\hskip-.1in &  \hskip -.1in\mat 0 &  \hskip -.1in\ddots&  \hskip-.1in\mat 0\\
\mat 0 & \hskip-.1in\cdots & \hskip -.1in &  \hskip -.1in\mat 0& \hskip-.1in{\mat Z}\mode{S[c]}
\end{array}\hskip -.075in\right] %\mat G \mat G\inv{}
\hskip -.025in
\underbrace{\left[\hskip -.05in
\begin{array}{c}
%{\mat U}\mode {c,1} \mat T\m
\left({\mat U}\mode {C,1} 
...%\dots 
\otimes {\mat U}\mode {(c+1),1} \otimes {\mat U}\mode {(c-1),1} 
...%\dots 
\otimes {\mat U}\mode {0,1}\right)\tp\\
\vdots\\
%{\mat U}\mode {c,s} \mat T\mode{s[c]} 
\left({\mat U}\mode{C,s} 
...%\dots
\otimes {\mat U}\mode{(c+1),s} \otimes {\mat U}\mode{(c-1),s} %...%
...%\dots
\otimes {\mat U}\mode{0,s}\right)\tp\\
\vdots \\
%{\mat U}\mode{c,S}\mat T\mode{S[c]}
\left({\mat U}\mode{C,S} 
...%\dots 
\otimes {\mat U}\mode{(c+1),S} \otimes {\mat U}\mode{(c-1),S} 
...%\dots
\otimes {\mat U}\mode{0,S}\right)\tp\\
\end{array} \hskip -.075in \right]}%\nonumber\\
\hskip-.075in\label{eq:independent_efficient}
\nonumber\\
&&
\hskip+1.1in{%\smaller
\left({\mat U}\mode {Cx} %\otimes 
\dots
\odot {\mat U}\mode {(c+1)x} \odot {\mat U}\mode {(c-1)x} %\otimes 
\dots 
\odot {\mat U}\mode {0x}\right)\tp}
\nonumber\\
&\hskip-.1in = %&
{\mat U}\mode {cx} \mat W\mode c\tp,\hskip-.35in
\vspace{-.5in}
\end{eqnarray}
%\end{align*}
%\vspace{-.2in}
%\end{strip}
%\vspace{-.05in}
%\vspace{\belowdisplayskip}\hfill\rule[-6pt]{0.4pt}{6.4pt}%
%\rule{\dimexpr(0.5\textwidth-0.5\columnsep-1pt)}{0.4pt}
% where $\mat W\mode m = \mat U\mode M \odot \dots \odot \mat
% U\mode {m+1} \odot \mat U\mode {m-1} \odot \dots \odot \mat U\mode 1$.
%\iffalse
%\vspace{-.25in}
%\\
\vspace{-.2in}

\noindent
where $\otimes$ is the Kronecker product,\footnote{The Kronecker product of $\mat U \in \Reals^{I×J}$ and $\mat V \in \Reals^{K×L}$ is the $IK \times JL$ matrix defined as $[\mat U \otimes \mat V ]_{ik,jl} = u_{ij}v_{kl}$.}and $\odot$ is the block-matrix Kahtri-Rao product.\footnote{The Khatri-Rao product of $\left[\mat U\mode1 \dots \mat U\mode n \dots \mat U\mode N\right]\odot \left[\mat V\mode1 \dots \mat V\mode n \dots \mat V\mode N\right]$ with $\mat U\mode l \in \Reals^{I\times N\mode l}$ and $\mat V\mode l
\in \Reals^{K\times N\mode l}$ 
%is denoted as $\mat U \odot \mat V$ and its
%entries are computed by $[\mat U \odot \mat V]_{ik,l}=u_{il}v_{kl}$.
is a {\it block-matrix Kronecker product}; therefore, it 
can be
expressed as $\mat U \odot \mat V = [(\mat U\mode{1}\otimes \mat V\mode{1}) \dots (\mat U\mode{(l)}\otimes \mat V\mode{(l)}) \dots (\mat
U\mode{(L)}\otimes \mat V\mode{(L)})]$ \cite{Rao71}.} 
%\fi
The matricized block diagonal form of ${\ten Z}\tmode H$ in eq.~\ref{eq:Dapprox_matrix2} becomes evident when employing our modified data centric matrixizing operator  based on the defintion~\ref{def:matrixizing}, where the initial mode is the measurement mode.

The compositional hierarchical tensor factorization algorithm computes the mode matrix, ${\mat U}\mode{cx}$, by computing the %finding a local 
minimum of $e=\|\ten D
- \tilde{\ten Z}\tmode H \times\mode 0 \tilde{\mat U}\mode {0x} \dots \times\mode C \tilde{\mat U}\mode {Cx}\|^2$ by cycling
through the modes, solving for $\tilde{\mat U}\mode {cx}$ in the equation $\partial
e/\partial \mat U\mode {cx}=0$ while holding the core tensor $\ten Z\tmode H$ and all the other mode matrices
constant, and repeating until convergence. Note that
%\vspace{-.2in}
\begin{equation}
\hspace{-.115in}
{\partial e\over\partial \mat U\mode {cx}}\hskip-.025in=\hskip-.025in{\partial\over\partial \mat U\mode {cx}}
\|{\matize D c} - \mat U\mode {cx}\mat W\mode c\tp\|^2 \hskip-.025in=-{\matize D c}
\mat W\mode c + \mat U\mode {cx} \mat W\mode c\tp \mat W\mode c.
%\hskip+.015in%\nonumber
%\vspace{-.05in}
\end{equation}

\noindent
Thus, $\partial e/\partial \mat U\mode {c}{}\pixels=0$ implies that

\vspace{-.15in}
%\rule{\dimexpr(.5\textwidth-1\columnsep-0.4pt)}{0.4pt}%
%\rule{0.4pt}{6pt}
%\vspace{-.2in}
%\begin{strip}
%\vspace{-.2in}
%\begin{align*}
%\vspace{-1in}
\begin{eqnarray}
%\vspace{-1in}
&\hskip-.3in
\mat U\mode {cx}
%\hskip-.05in
=& \hskip-.05in
{\matize D c} \mat W\mode c \left(\mat W\mode c\tp \mat W\mode c\right)^{-1} %\\
%&=&{\hskip -.1in} 
={\matize D c} \mat W\mode c\tp{}\pinv{}%\hskip-.5in \\
%&{\hskip -.125in}=
\hskip-.75in
\\
&\hskip-.5in
=&\hskip-.285in 
{\matize D c} \left(\mat Z\tmode H{}\mode {[c]}\left({\mat U}\mode {Cx} \otimes 
...%\dots
%\otimes
{\mat U}\mode {(c+1)x} \otimes {\mat U}\mode {(c-1)x} \otimes 
...%\dots 
%\otimes 
{\mat U}\mode {0x}\right)\tp\right)\pinv{}%\hskip+.5in
\hskip-.75in\\% \nonumber \\ 
%&{\hskip -.125in}
&\hskip-.5in= &\hskip-.285in
{\matize D c}\hskip-.025in
\left({\mat U}\mode {Cx} %\otimes 
\hskip-.05in\odot
...%\dots
{\mat U}\mode {(c+1)x} \hskip-.025in\odot \hskip-.025in{\mat U}\mode {(c-1)x} 
\hskip-.025in\odot
...%\dots 
%\odot 
{\mat U}\mode {0x}\right)\hskip-.05in\tp{}\pinv{}%\nonumber\\
%&&
\hskip-.025in\left[\hskip -.075in
\begin{array}{ccccc}
{\mat Z}\mode{0[c]}\pinv{} &\hskip-.1in\mat 0& \hskip -.1in&  \hskip -.1in\cdots&\hskip-.1in\mat 0 \\
\mat 0 &\hskip-.1in\ddots&  \hskip -.1in\mat 0&  \hskip -.1in &\hskip-.1in\vdots  \\
\vdots  &\hskip-.1in\mat 0 & \hskip -.1in{\mat Z}\mode{s[c]}\pinv{}& \hskip -.1in \mat 0 & \hskip-.1in \\
 &\hskip-.1in &  \hskip -.1in\mat 0 &  \hskip -.1in\ddots&  \hskip-.1in\mat 0\\
\mat 0 & \hskip-.1in\cdots & \hskip -.1in &  \hskip -.1in\mat 0& \hskip-.1in{\mat Z}\mode{S[c]}\pinv{}
\end{array}\hskip -.075in\right]
\end{eqnarray}
%\end{align*}
%\vspace{-.2in}
%\end{strip}
%\vspace{-.05in}
%\vspace{\belowdisplayskip}\hfill\rule[-6pt]{0.4pt}{6.4pt}%
%\rule{\dimexpr(0.5\textwidth-0.5\columnsep-1pt)}{0.4pt}
%\vspace{-.05in}
\noindent
whose $\mat U\mode{c,s}$ sub-matrices are then subject to orthonormality constraints.   

Solving for the optimal core tensor, $\ten Z\tmode H$, the data tensor, $\ten D$, approximation is expressed in vector form as,% $\vec d=\mbox{vec}(\ten D)$, form as
\vspace{-.05in}
\begin{eqnarray}
e=\|\mathrm{vec}({\ten D}) - (\tilde{\mat U}\mode{Cx} \otimes \dots \otimes\tilde{\mat U}\mode{cx}\otimes \dots\otimes \tilde{\mat U}\mode{0x})\mathrm{vec}(\tilde{\ten Z}\tmode H)\|.
\vspace{-.1in}
\end{eqnarray}

\vspace{-.05in}
\noindent
Solve for the non-zero(nz) terms of $\ten Z\tmode {H}$ in the equation $\partial e/\partial ({\small\ten Z\tmode H}) =0$,
by removing the corresponding zero columns of the first matrix on right side of the equation below, performing the pseudo-inverse, %while holding all mode matrices constant, 
and setting
%\vskip -.05in
\begin{equation}
%\vspace{-.15in}
\mathrm{vec}(\ten Z\tmode H)\mode{nz}=(\mat U\mode{Cx} \otimes \dots \otimes\mat U\mode{cx}\otimes \dots\otimes \mat U\mode{0x})\pinv{}\mode{nz}\mathrm{vec}(\ten D).
\vspace{-.05in}
\label{eq:tensor_model}
\end{equation}
%and 
%\begin{wrapfigure}{r}{.525\textwidth}
%\vspace{-.375in}
%\hspace{.15in}
%\begin{minipage}{1\linewidth}
%\hskip +0in 
%\begin{algorithm}
%\hrule
%\medskip
%% \underline{\bf Compositional $M$-Mode SVD Algorithm:}
%\vskip +.2in
%\begin{minipage}{\textwidth}
\begin{algorithm}[!phb]
%\begin{minipage}
%\begin{algorithm}
%\hrule
%\medskip
%% \underline{\bf Compositional $M$-Mode SVD Algorithm:}
%\vskip -.2in
\begin{minipage}{\linewidth}
%\vskip -.2in
{\bf Input:} Data tensor, $\ten D\in\Reals^{I\mode 0\times I\mode 1\times\dots\times I\mode C}$, filters $\mat H\mode s$, %data tensor tree parameterization, $t$, 
and desired dimensionality reduction $\tilde{J}\mode 1, \dots, \tilde{J}\mode C$.
%\end{minipage}
%\hskip -.5in%
%\begin{minipage}{\linewidth}
\begin{enumerate}
\item[]
\hskip -.3in {1.\sl Initialization:}%\\
%\hskip-.3in
\item[1a.] 
Decompose each data tensor segment, $\ten D\mode s=\ten D\times \mat H\mode s$, by employing the M-mode SVD.  
$$\ten D\mode s= \ten Z \times\measure \mat U\measure{}\mode{,s}\dots\times\mode c \mat U\mode{c,s}\dots\times \mode C\mat U\mode{C,s}$$
\item[1b.]For $c = 0,1,\dots,C$, set $\mat U\mode{cx}=[\mat U\mode{c,1} ... \mat U\mode{c,s} ... \mat U\mode{c,S}]$, and truncate to $\tilde J\mode c$ columns %the mode matrix $\mat U\mode {cx}\in\Reals^{I\mode c\times\tilde J\mode c}$ 
%by 
 by sorting all the eigenvalues from all data segments and 
%appropriately 
deleting the columns corresponding to the lowest eigenvalues from various $\mat U\mode {c,s}$ and the rows from $\mat Z\mode {s[c]}$ 
\iffalse
deleting the columns from various $\mat U\mode {c,s}$ and rows from $\mat Z\mode {s[c]}$ % accordingly.% based on 
corresponding to the lowest $(S I\mode c - \tilde J\mode c)$ eigenvalues.
% truncate to $\tilde J\mode c$ mode matrix $\mat U\mode{cx} \in \Reals^{I_c \times \tilde{J}_{c}}$ by appropriately truncating the mode matrices $\mat U\mode{c,s}$ based on the lowest $(SI\mode c − \tilde{J}\mode c)$ eigenvalues across all data segments.
\fi
\iffalse
\begin{itemize}
\item[]Determine index order of eigenvalues for every compositional causal mode matrix,  %$[\sigma\mode{11,1} & \dots & {\sigma \mode {i_ci_c,1}}  & \dots& {\sigma\mode{I_cI_c,1}} | \dots \sigma\mode{11,s} \dots \sigma\mode{i_ci_c,s}  \dots \sigma\mode{I_cI_c,s}  \dots \sigma\mode{11,S}  \dots \sigma\mode{i_c i_c,S} \dots \sigma\mode{I_c I_c,S}]$:
\item[]Compute permutation matrix, $\mat G$
\item[]Construct compositional truncated mode matrices, $\mat U\mode{cx}$
\end{itemize}
\fi
\item[]
%\label{step:optimization}
{\vskip+.05in\hskip -.3in 2.\sl Optimization via alternating least squares:}\\
%Set $e\mode 0 \assign \infty$.\\
\item[]
{\vskip-.15in \hskip -.275in}Iterate for $n \assign 1,\dots,N$
\begin{itemize}
\item[]
%\hskip -.15in
{\hskip -.325in}
For $c \assign 0,\dots,C$, 
\item[]
\hskip-.3in 2a. Compute mode matrix $\mat U\mode {cx}$ while holding the rest fixed. %according to
%\begin{itemize}
%\item[] 
%Set 
\begin{eqnarray}
\hskip+.25in
\mat U\mode{cx} 
&\hskip-.15in\assign\hskip-.125in&
{\matize D c}\hskip-.055in
\left({\mat U}\mode {Cx} %\otimes 
\hskip-.05in\odot
...%\dots
{\mat U}\mode {(c+1)x} \hskip-.025in\odot \hskip-.025in{\mat U}\mode {(c-1)x} 
\hskip-.025in\odot
...%\dots 
%\odot 
{\mat U}\mode {0x}\right)\hskip-.055in\tp{\hskip-.025in}\pinv{}
%\nonumber\\
%& &
\hskip-.025in
\left[\hskip -.075in
\begin{array}{ccccc}
{\mat Z}\mode{0[c]}\pinv{} &\hskip-.1in\mat 0& \hskip -.1in&  \hskip -.1in\cdots&\hskip-.1in\mat 0 \\
\mat 0 &\hskip-.1in\ddots&  \hskip -.1in\mat 0&  \hskip -.1in &\hskip-.1in\vdots  \\
\vdots  &\hskip-.1in\mat 0 & \hskip -.1in{\mat Z}\mode{s[c]}\pinv{}& \hskip -.1in \mat 0 & \hskip-.1in \\
 &\hskip-.1in &  \hskip -.1in\mat 0 &  \hskip -.1in\ddots&  \hskip-.1in\mat 0\\
\mat 0 & \hskip-.1in\cdots & \hskip -.1in &  \hskip -.1in\mat 0& \hskip-.1in{\mat Z}\mode{S[c]}\pinv{}\nonumber
\end{array}\hskip -.075in\right]
\end{eqnarray}

\item[] 
\hskip -.15in
Set $\hat{\mat U}\mode {c,s}$ to the $\tilde R\mode {c,s}$ leading left-singular
vectors of the SVD\footnote{\label{note:gram-schmidt}The complexity of computing
the SVD of an $m\times n$ matrix $\mat A$ is $O(mn\min(m,n))$, which
is costly when both $m$ and $n$ are large. However, we can efficiently
compute the $\tilde R$ leading left-singular vectors of $\mat A$ by
first computing the rank-$\tilde R$ modified Gram Schmidt (MGS)
orthogonal decomposition $\mat A \approx \mat Q\mat R$, where $\mat Q$
is $m\times \tilde R$ and $\mat R$ is $\tilde R\times n$, and then
computing the SVD of $\mat R$ and multiplying it as follows: $\mat A\approx \mat Q(\tilde{\mat U}\mat S%\Sigma
\mat V\tp) = (\mat Q\tilde{\mat U})\mat S%\Sigma 
\mat V\tp  = \mat U\mat S%\Sigma
\mat V\tp$.
% In practice we discard all factors whose singular values when
% normalized by the largest value are less than some desired accuracy
% ($> \epsilon$).
} of 
\item[] 
\hskip -.15in$\mat U\mode{c,s}$, a subset of the columns in $\mat U\mode{cx}$.Update $\mat U\mode{c,s}$ in ${\mat U}\mode{cx}$ with $\hat{\mat U}\mode{c,s}$
%\end{itemize}

%$\ten D\mode{s}=\ten T\mode{s} \times\mode {2} \mat U\mode {2,2} \times \dots\times\mode {m} \mat U\mode {m,s} \dots \times\mode {M} \mat U\mode {m,S}$
\item[]
\hskip-.3in2b. Set the non-zero(nz) entries of $\mathrm{vec}(\ten Z\tmode H)\mode{nz}$  based on:%according to eq.~\ref{eq:extcore_general}\\
%\vspace{-.05in}
\begin{eqnarray*}
\hskip+.3in\mathrm{vec}(\ten Z\tmode H)\mode{nz}=(\mat U\mode{Cx} \otimes \dots \otimes\mat U\mode{cx}\otimes \dots\otimes \mat U\mode{0x})\pinv{}\mode{nz}\mathrm{vec}(\ten D).
%\mathrm{vec} ( \ten Z{\tmode H} )\mode {nz}
%\hskip -.1in &=& \hskip -.1in
%\mathrm{vec} (
%\ten D \times\mode 0 
%\mat U\mode {0x}\pinv \dots 
%%\times\mode c \mat U\mode {cx}\pinv 
%%\dots 
%\times\mode C 
%\mat U\mode {Cx}\pinv{} 
%)\mode {nz}\nonumber
\end{eqnarray*}
%$\ten T\mode{s} \assign \ten D\mode{s} \times\mode 2 \mat U\mode {2,s}\pinv{} \dots\times\mode m \mat U\mode {m,t}\pinv{} \dots \times\mode M \mat U\mode {M,t} \pinv{} $.
\end{itemize}
\hskip-.275in until convergence.
~\footnote{Note that $N$ is a pre-specified maximum number of iterations. A possible convergence criterion is to compute at each iteration the approximation error $e\mode n \assign \| \ten D - \tilde{\ten D} \|^2$ and test if $e\mode {n-1} - e\mode n\le\epsilon$ for sufficiently small tolerance $\epsilon$.
\label{note:convergence-criterion}}
%\footnote{See Footnote~\ref{note:convergence-criterion} in Algorithm~\ref{alg:cp}.}
\end{enumerate}
%\end{minipage}%\footnote{See
%Footnote~\ref{note:convergence-criterion} in Algorithm~\ref{alg:cp}.}

{\bf Output} converged matrices ${\mat U}\mode{1x},\dots,{\mat U}\mode{Cx}$ and tensor ${\ten Z}{\tmode H}$.
%\vspace{-.2in}
\medskip
\hrule
\vspace{-.075in}
\end{minipage}
\caption[Compositional Hierarchical Tensor Decomposition algorithm]{Compositional Hierarchical Tensor Factorization.}
\label{alg:compositional-m-mode-svd}
\end{algorithm}
%\vspace{-.25in}
%\end{minipage}
%\hrule
%\vspace{-1.75in}
%\end{wrapfigure}
%

\vskip-.05in
\noindent
Repeat all steps until convergence.This optimization is the basis of the Compositional Hierarchical Tensor Factorization, Algorithm~\ref{alg:compositional-m-mode-svd}.
%This algorithm is able to handle a fully compositional or partially compositional factorization as
\\
\vspace{-.05in}
\noindent
\underline{Completely Overlapping parts:} When the data tensor is a collection of overlapping parts that have the same multilinear-rank reduction, Fig.~\ref{fig:fully_compositional}e,  %as in the case of block-tensor decomposition, 
the extended-core data tensor, $\ten T\tmode H$, computation in matrix form reduces to eq.~\ref{eq:overlapping_efficient}
%containing disparate pieces of information, as in the case of an object whose overlapping parts are the red/green/blue channels, or the overlapping parts are the levels of a Laplacian, 
\\
%\iffalse
\vspace{-.2in}
\begin{eqnarray}
%\hspace{-.35in}
\hskip-.55in
{\ten D}
&\hskip-.05in =& \hskip-.05in
{\ten T}\tmode H\times\mode 1 {\mat U}\mode{1x}\dots \times\mode c {\mat U}\mode{cx}\dots \times\mode C {\mat U}\mode{Cx}
\\
\vspace{+.2in}
\hspace{-.35in}
%hskip -1.5in 
{\mat D}\mode{[0]}
&\hskip -.05in =& \hskip -.05in
%&=&
%{\mat U}\mode {cx} 
%\underbrace{
{\mat T}\tmode H{}\mode{[0]}
(
{\mat U}\mode {Cx} 
\dots 
%\otimes%\odot 
%{\mat U}\mode {(c+1)x} 
\otimes%\odot 
{\mat U}\mode {cx}
\dots 
\otimes%\odot
{\mat U}\mode {1x}
)\tp \\%}\mode{\mat W\tp} \\%\nonumber
\nonumber\\
% comment
%\hskip-.5in =\hskip -.15in 
\vspace{-.1in}
\left[\begin{array}{c}
{\mat T}\mode{1[0]}\\
\vdots\\
{\mat T}\mode{s[0]}\\
\vdots\\
{\mat T}\mode{S[0]}
\end{array}
\right]\tp
&\hskip -.15in =& \hskip -.1in
%&=& %\hskip -.15in
{\mat D}\mode {[0]}%\hskip-.05in
\underbrace{\left[\begin{array}{c}
%\hskip-.1in
%\mat Z\mode {1[c]} 
({\mat U}\mode {C,1} \dots 
%\otimes {\mat U}\mode {c+1,1} 
\otimes {\mat U}\mode {c,1} \dots 
\otimes {\mat U}\mode {1,1})\tp\\
\vdots\\
%\hskip-.1in
%\mat Z\mode {s[c]}
({\mat U}\mode {C,s} \dots 
%\otimes {\mat U}\mode {c+1,s} 
\otimes {\mat U}\mode {c,s} \dots 
\otimes {\mat U}\mode {1,s})\tp\\
\vdots\\
%\hskip-.1in
%\mat Z\mode {S[c]}
({\mat U}\mode {C,S} \dots 
%\otimes {\mat U}\mode {c+1,S} 
\otimes {\mat U}\mode {c,S} \dots
\otimes {\mat U}\mode {1,S})\tp
\end{array}%\hskip-.1in
\right]\pinv{}}_{{\left(\begin{array}{c}
%\hspace{+.025in}
{\mat U}\mode {Cx} 
\cdots 
%\odot {\mat U}\mode {c+1x} 
\odot {\mat U}\mode {cx} 
\cdots
\odot {\mat U}\mode {1x}
\end{array}
\right)\tp}\pinv{}}. 
%\hskip-.2in
\label{eq:overlapping_efficient}
% \hskip -.5in 
%\hskip -.15in
\label{eq:U_overlapping}
%\vspace{-.05in}
\end{eqnarray}
\vspace{-.125in}

\noindent
\underline{Independent Parts:} When the data tensor is a collection of observations made up of non-overlapping parts, Fig.~\ref{fig:independentparts}d, the data tensor decomposition
%, eq.~\ref{eq:DinDF}, 
reduces to the concatenation of a $M$-mode SVD of individual parts,  
\vspace{-.05in}
\begin{eqnarray}
\hskip -.2in 
\left[%\hskip -.05in
\begin{array}{c}
\mat D\mode{1[c]}\\
\vdots\\
\mat D\mode {s[c]}\\
\vdots\\
\mat D\mode {S[c]}\\
\end{array}% \hskip -.05in 
\right]
%\hskip -.15in&=&\hskip -.2in 
&\hskip-.075in=%\approx
&\hskip-.075in
\left[\hskip -.05in
\begin{array}{ccccc}
{\mat U}\mode {c,1} \mat Z\mode{0[c]} ({\mat U}\mode {C,1} 
\dots 
\otimes {\mat U}\mode {(c+1),1} \otimes {\mat U}\mode {(c-1),1} 
\dots 
\otimes {\mat U}\mode {0,1})\tp\\
\vdots\\
{\mat U}\mode {c,s} \mat Z\mode{s[c]} ({\mat U}\mode{C,s} 
\dots
\otimes {\mat U}\mode{(c+1),s} \otimes {\mat U}\mode{(c-1),s} %...%
\dots
\otimes {\mat U}\mode{0,s})\tp\\
\vdots \\
{\mat U}\mode{c,S}\mat Z\mode{S[c]}({\mat U}\mode{C,S} 
\dots 
\otimes {\mat U}\mode{(c+1),S} \otimes {\mat U}\mode{(c-1),S} 
\dots
\otimes {\mat U}\mode{0,S})\tp \\
\end{array} \hskip -.05in \right]
% comment
%~~~\hskip -.1in & &\hskip -.1in~({\mat U}\mode{Cx} \dots \odot 
%{\mat U}\mode{(c+1)x} \odot {\mat U}\mode{(c-1)x} \dots \odot \tilde{\mat U}\mode{1x})\tp
%\left[\begin{array}{c}
%\tilde{\mat D}\mode{\mbox{\tiny F[{\it m,1}]}} \\
%\vdots \\
%\tilde{\mat D}\mode{\mbox{\tiny F[{\it m,s}]}} \\
%\vdots\\
% \tilde{\mat D}\mode{\mbox{\tiny F[{\it m,S}]}}
%\end{array}\right]\tp \\
%&=&
%\left[\begin{array}{c}
%\tilde{\mat U}\mode{m,1}\matize T {m,1}(\tilde{\mat U}\mode{M,1} \dots \otimes %\tilde{\mat U}\mode{m+1,1} \otimes \tilde{\mat U}\mode{m-1,1} \dots \otimes \tilde{\mat U}\mode{2,1})\tp\\
%\vdots\\
%\tilde{\mat U}\mode{m,s}\matize T {m,s}(\tilde{\mat U}\mode{M,s} \dots \otimes \tilde{\mat U}\mode{m+1,s} \otimes \tilde{\mat U}\mode{m-1,s} \dots \otimes \tilde{\mat U}\mode{2,s})\tp\\
%\vdots\\
%\tilde{\mat U}\mode{m,S}\matize T {m,S}(\tilde{\mat U}\mode{M,S} \dots \otimes \tilde{\mat U}\mode{m+1,S} \otimes \tilde{\mat U}\mode{m-1,S} \dots \otimes \tilde{\mat U}\mode{2,S})\tp
%\end{array}\right]\tp\\
\end{eqnarray}
%\end{minipage}
%\end{wrapfigure}
Note, that every $\mat D\mode{s[c]}$ row does not contain any terms from any other segment-part except segment $s$.  Thus,  every
$\mat U\mode {c,s}$ and $\ten Z\mode s$ are computed by performing multilinear subspace learning, Algorithm~\ref{alg:m-mode-svd}, on the $\ten D\mode s$ and the results are appropriately concatenated in $\tilde{\mat U}\mode {cx}$ and $\tilde{\ten Z}\tmode H$.% When performing dimensionality reduction for a particular causal factor, $c$, the singular values across data segments and dimensionality reduction is performed by .  %
\iffalse
\begin{figure}%{R}{.55\textwidth}%[!bthp]
%\vspace{-.355in}
\hspace{+.0in}
%\begin{center}%\
%\centerline{
%\fbox{\rule{0pt}{2in} \rule{.9\linewidth}{0pt}}
\includegraphics[width=.7\textwidth]{images/Compositional-TensorFaces-13-01.png}
%}

%\centerline{(a)}
%\centerline{
%\makebox[4in]
%{\includegraphics[width=.02\linewidth]{images/arrow.png}}
%\hfill}
%\centerline{
%\includegraphics[height=.45\linewidth]{images/Hierarchical_subdivision_hierarchical_tensorfaces_gaussian_pyramid13only-01.png}
%\hfill
%\makebox[2in]{ 
%\includegraphics[height=.47\linewidth]{images/deepface4.png}}
%\hfill
%}
%\centerline{\hfill \makebox[3in]{(b)} \makebox[4in]{(c)}}
%\end{center}    
\caption{Compositional Hierarchical TensorFaces learns a hierarchy of features and represents each person as a compositional representation of parts. Figure depicts the training data factorization, \\$\ten D = \ten T{\tmode H} \times\illums \mat U \illums \times\views \mat U\views \times\people \mat U\people$, where an observation is represented as $\vec d(\vec p,\vec v,{\vec l}) = \ten T{\tmode H} \times\illums \vec l\tp \times\views \vec v\tp \times\people \vec p\tp $.%}
%(b) depicts an architecture that shows the data flow that parallels 
%(c) CNN architecture %and hierarchy of features parallels 
%(Image is a modified RSIP image). %Often, deep neural networks are not interpretable. For example, the filters of VGG-Face, the most accurate face verification on LFW, do not have an easy or intuitive human interpretation ~\cite{parkhi2015}
}
\label{fig:HTensorFaces}
\vspace{-.2in}
\end{figure}
\fi

\begin{figure*}[!t]%htbph]
\vspace{-.25in}
%\hspace{.2in}
\centerline{
\hfill
\begin{tabular}{c}
\includegraphics[width=.74\linewidth]{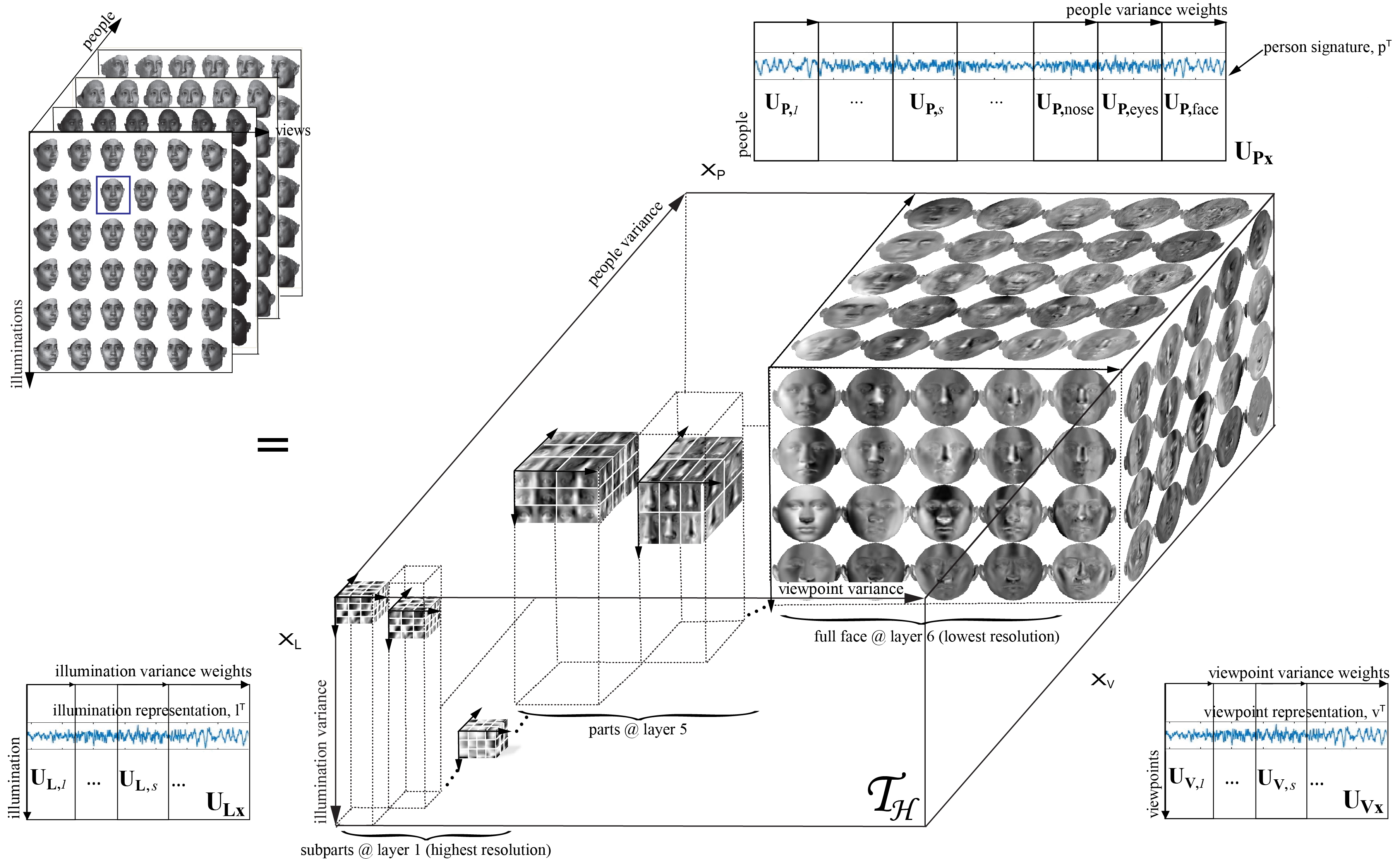}
\end{tabular}
\hfill
\begin{tabular}{l}
\includegraphics[width=.19\textwidth]{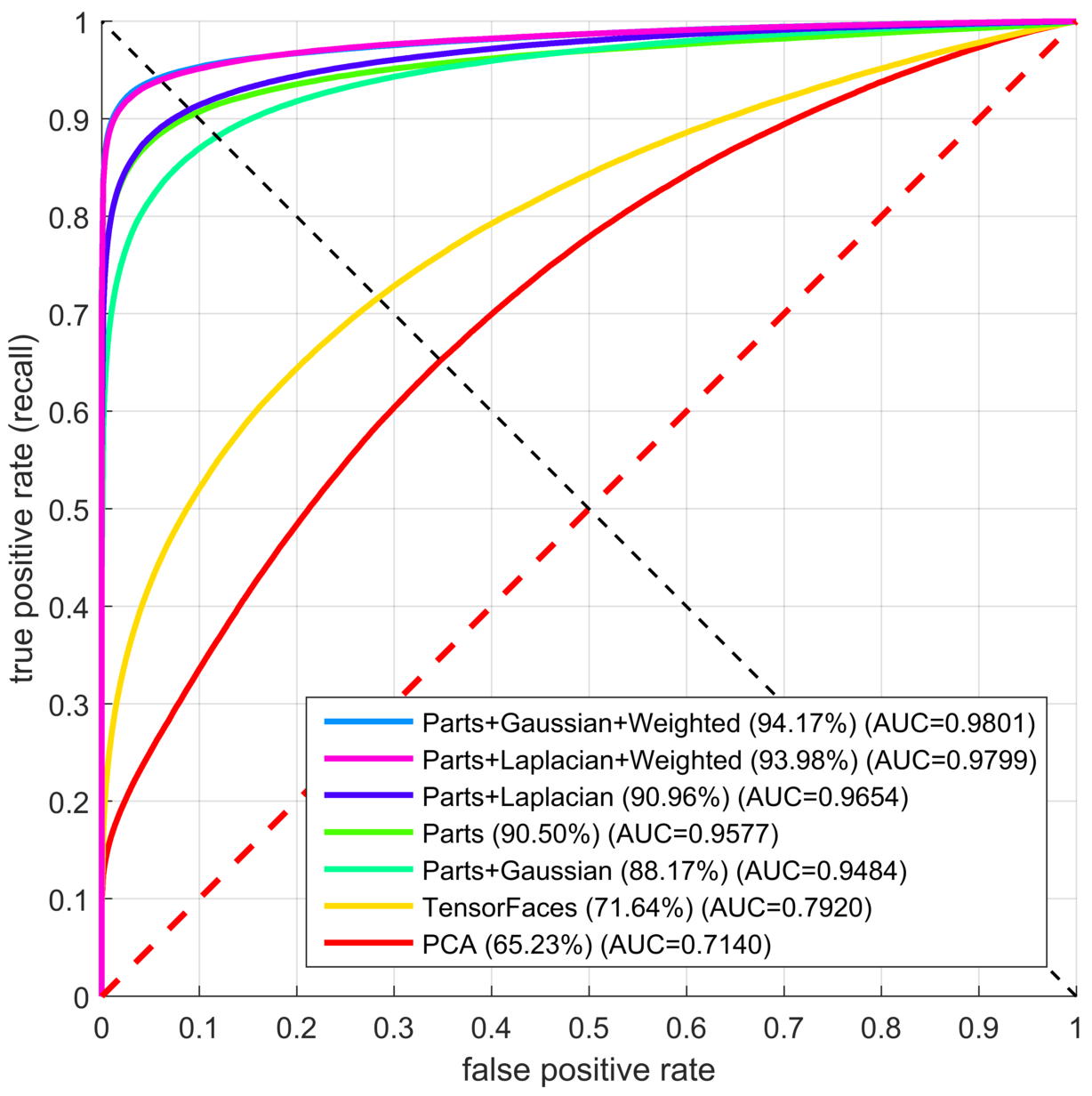} \\ 
\vspace{-.1in}\hskip +.7in{\footnotesize (b)}
\vspace{+.2in}
\\ 
\\
\includegraphics[width=.19\textwidth]{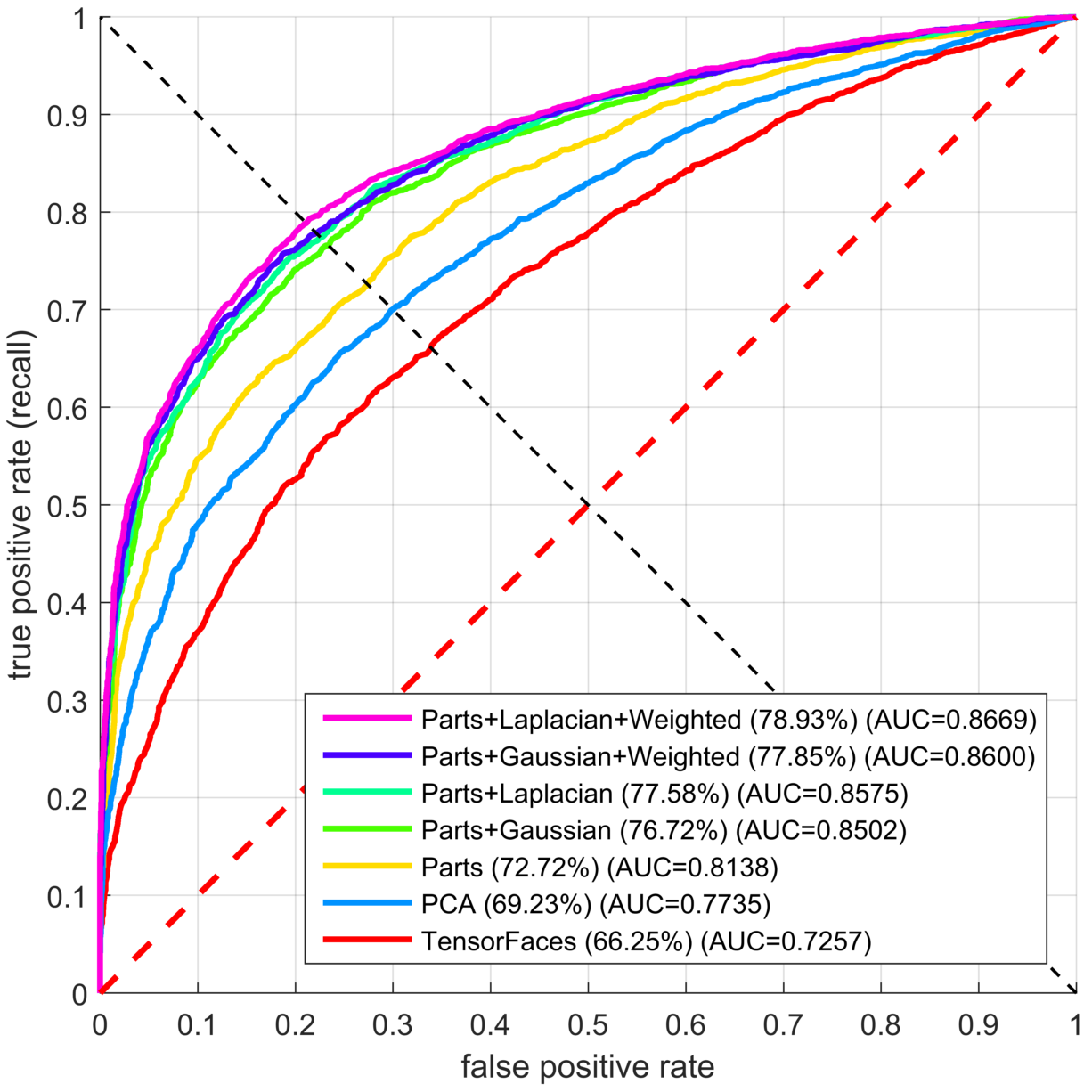}\\
%\vspace{-.05in}
{\hskip +.7in\vspace{-.0in}{\footnotesize (c)}}\\
\end{tabular}
\hfill
}
\vspace{-.15in}
\centerline{\footnotesize \hspace{2.5in} (a)\hfill}
\vspace{-.17in}
  \caption{a) Compositional Hierarchical TensorFaces learns a hierarchy of features and represents each person as a part-based compositional representation. Figure depicts the training data factorization, $\ten D = \ten T{\tmode H} \times\illums \mat U \illums \times\views \mat U\views \times\people \mat U\people$, where an observation is represented as $\vec d(\vec p,\vec v,{\vec l}) = \ten T{\tmode H} \times\illums \vec l\tp \times\views \vec v\tp \times\people \vec p\tp $. b) ROC curves for the University of Freiburg 3D Morphable Faces dataset. c) ROC curves for the LFW dataset.
The average accuracies are listed next to each method, along with the area under the curve (\emph{AUC}).
\emph{Parts} refers to using Compositional Hierarchical TensorFaces models to separately analyze facial parts.
\emph{Gaussian}, \emph{Laplacian} refers to using Compositional Hierarchical TensorFaces on a Gaussian/Laplacian data pyramid.
%to analyze facial parts at different scales.
}
  %\label{fig:tensor}
  \label{fig:HTensorFaces}
  \vspace{+.1in}
\end{figure*}
\begin{table*}[!t]%phbth]%{.9\textwidth}%[!bht]
\vspace{-.265in}
\begin{center}
\begin{tabular}{|c|c|c|c|c|c|c|c|}
\hline
%\multirow{2}{*}
{\bf \begin{tabular}[c]{@{}l@{}} Test \\Dataset\end{tabular}}  & %\multirow{2}{*}[+1ex]
{\bf PCA}  & 
%\multirow{2}{*}[+1ex]
{\bf TensorFaces}  & 
\multicolumn{5}
{c|}{\bf Compositional Hierarchical TensorFaces
%\Tstrut\Bstrut
}       \\ %\cline{4-8}
& & & \begin{tabular}[c]{@{}c@{}}Pixels\end{tabular} & \begin{tabular}[c]{@{}c@{}}Gaussian\\Pyramid\end{tabular}  & \begin{tabular}[c]{@{}c@{}}Weighted\\ Gaussian\\Pyramid\end{tabular} & \begin{tabular}[c]{@{}c@{}}Laplacian\\Pyramid\end{tabular} & \begin{tabular}[c]{@{}c@{}}Weighted\\Laplacian\\Pyramid\end{tabular}  
\\ 
\hline 
\hline 
% &  & & & & & &   \cr
Freiburg  &  65.23\%   &  71.64\%  &  90.50\%  &  88.17\%  & 94.17\%  & 90.96\%  &  93.98\% \cr
% &  & & & & & & \\
\hline
 %&  & & & & & & \cr
 \begin{tabular}[c]{@{}l@{}} LFW\\ \ \end{tabular} & 
 \begin{tabular}[c]{@{}c@{}}\ 69.23\%\Tstrut\\ \rpm 1.51\end{tabular}  &  \begin{tabular}[c]{@{}c@{}}66.25\%\Tstrut\\ \rpm 1.60\end{tabular} &  \begin{tabular}[c]{@{}c@{}}72.72\%\Tstrut\\ \rpm 2.14\end{tabular} & \begin{tabular}[c]{@{}c@{}}76.72\%\Tstrut\\ \rpm 1.65\end{tabular}  & \begin{tabular}[c]{@{}c@{}}77.85\%\Tstrut\\ \rpm 1.83\end{tabular} & \begin{tabular}[c]{@{}c@{}}77.58\%\Tstrut\\ \rpm 1.45\end{tabular}   & \begin{tabular}[c]{@{}c@{}}78.93\%\Tstrut\\ \rpm 1.77\end{tabular}  \cr
 %&  & & & & & & \cr
\hline
\end{tabular}
%\vspace{.025in}
\caption[Empirical results reported on two facial image datasets]{Empirical results reported for LFW : PCA, TensorFaces and Compositional Hierarchical TensorFaces.
\emph{Pixels} denotes indepent facial part analysis%zed separately, but without any multiresolution pyramid.
\emph{Gaussian}/\emph{Laplacian} use a multiresolution pyramid to analyze facial features at different scales.
\emph{Weighted} denotes a weighted nearest neighbhor.% composite signature.
\newline \underline{Freiburg Experiment:}
\newline 
Train on Freiburg: 6 views (\protect\rpm 60\ensuremath{^\circ},\protect\rpm 30\ensuremath{^\circ},\protect\rpm 5\ensuremath{^\circ}); 6 illuminations (\protect\rpm 60\ensuremath{^\circ},\protect\rpm 30\ensuremath{^\circ},\protect\rpm 5\ensuremath{^\circ}), 45 people
\newline 
Test on Freiburg:\mbox{\hspace{+.1in}} 9 views (\protect\rpm 50\ensuremath{^\circ}, \protect\rpm 40\ensuremath{^\circ}, \protect\rpm 20\ensuremath{^\circ}, \protect\rpm 10\ensuremath{^\circ}, 0\ensuremath{^\circ}), 9 illums (\protect\rpm 50\ensuremath{^\circ}, \protect\rpm 40\ensuremath{^\circ}, \protect\rpm 20\ensuremath{^\circ}, \protect\rpm 10\ensuremath{^\circ}, 0\ensuremath{^\circ}), 45 different people
\newline \underline{LFW Experiment:} Models were trained on approximately half of one percent ($0.5\%<1\%$) of the $4.4$M images used to train DeepFace.
\newline 
Train on Freiburg: %100 people\\
\\
%\mbox{\hspace{+1in}}
15 views (\protect\rpm 60\ensuremath{^\circ},\protect\rpm 50\ensuremath{^\circ},
\protect\rpm 40\ensuremath{^\circ},\protect\rpm 30\ensuremath{^\circ}, 
\protect\rpm 20\ensuremath{^\circ}, \protect\rpm 10\ensuremath{^\circ},\protect\rpm 5\ensuremath{^\circ}, 0\ensuremath{^\circ}), 
15 illuminations 
(\rpm 60\ensuremath{^\circ},\protect\rpm 50\ensuremath{^\circ},
\protect\rpm 40\ensuremath{^\circ},\protect\rpm 30\ensuremath{^\circ}, 
\protect\rpm 20\ensuremath{^\circ}, \protect\rpm 10\ensuremath{^\circ},\protect\rpm 5\ensuremath{^\circ}, 0\ensuremath{^\circ}), 100 people
%\mbox{\hspace{+1in}} 
\newline Test on LFW: We report the mean accuracy and standard deviation across standard literature partitions~\cite{Huang2007}, 
%\mbox{\hspace{+.75in}}
following the \\{\em Unrestricted, labeled outside data} supervised protocol.}
\label{tab:results}
\end{center}
\vspace{-.3in}
\end{table*}

%\vskip +.15in\\
\vspace{-.1in}
\section{Compositional Hierarchical TensorFaces}
\label{sec:Compositional-TensorFaces}

\noindent
\underline{Training Data: } In our experiments, we employed gray-level facial training images rendered from 3D
scans % of the subject. The 100 face scans were 
of 100 subjects. The scans were recorded using a CyberwareTM
3030PS laser scanner and are part of the 3D morphable
faces database created at the University of Freiburg~\cite{Blanz99}.
%100 subjects. 
 Each subject was combinatorially imaged in Maya from 15 different viewpoints ($\theta = -60^\circ$ to $+60^\circ$ in $10^\circ$ steps on the horizontal
plane, $\phi=0^\circ$) 
with $15$ 
different illuminations (
$\theta = -35^\circ$
to 
$+35^\circ$ in $5^\circ$ increments on a plane inclined at 
$\phi = 45^\circ$). 
%Fig. 4(b)shows the full set of 225 images for one of the subjects
%with viewpoints arrayed horizontally and illuminations arrayed vertically. 
%The image set was rendered
\\
\iffalse
Our facial recognition training data consists of synthetically rendered %synthetically generated 
gray-level facial images of 100 subjects that were scanned using a CyberwareTM
3030PS laser scanner and are part of the 3D morphable
faces database created at the University of Freiburg

Synthethic images were trivially rendered %generated 
with mininmal cost from 15 different viewpoints
under 15 different illuminations ($\theta=−60^\circ$
to $+60^\circ$ in $10^\circ$ steps on an inclined plane $\phi=45^\circ)$. 
\fi

\vspace{-.176in}

\noindent
\underline{Data Preprocessing:}
%In order to compare like-with-like
%Global features are not robust against pose
%changes since global features are highly sensitivse to translation and rotation of
%the face. To avoid this problem, 
%images are warped to a template face. 
%The alignment step before analyzing and classifying a face. 
%Image correspondence were determined for a number of prominent points in the face like the eye corners, the nostrils, or the corners of the mouth by employing the facial feature detection algorithm of 
%In order to compare like-with-like, To align each i
%Image were brought into a canonical coordinate system by performing a piece-wise affine warp to a template shape.
%via a piecewise-affine warp. 
Facial images were warped  to an average face template 
by a piecewise affine transformation given a set of facial landmarks obtained by employing  Dlib software~\cite{dlib09,Kazemi2014,Si13}. %,hatamizadeh2019end}. %,hatamizadeh2019deep,ivanov09}. % %(Fig.~\ref{fig:warping}). 
Illumination was normalized with an adaptive contrast histogram equalization
algorithm, but rather than performing contrast correction on the entire image, 
subtiles of the image were contrast normalized, and tiling artifacts were eliminated through interpolation. Histogram clipping was employed to avoid over-saturated regions. Each image, $\vec d \in \Reals^{I\measure \times 1}$, %that has been pre-warped to a template
was convolved with a set of filters $\{\mat H\mode s  \|  s = 1 ... S \}$, %be a bank of filters.% 
and the filtered images, $\vec d \times\measure \mat H\mode s $,
%with that 
resulted either in a Gaussian or Laplacian image pyramid. 
% which capture edges and blobs.% Let $\{\mat H\mode k  \|  k = 1 \dots K \}$ be a bank of filters.% and $\mat H\mode k * \mat D$ be the filtered image. 
%%The filtered image is efficiently computed with a 
%by performing a simple matrix-vector multiplication, $\vec d\mode s = \mat H\mode s \vec d$ %when the image is vectorized, $\vec d$,  
%where the filter, $\mat H\mode s$ is a double circulant matrix.% and Let $\{\mat H\mode k  \|  k = 1 \dots K \}$ be a bank of filters.% and $\mat H\mode k * \mat D$ be the filtered image. 
% The filters may blurr, subsample and/or segment a facial feature when their spatial scope is limited.
Facial parts were segmented from the various layers.

\noindent
\underline{Experiments:} %The extended core $\ten T{\tmode H}$ that governs the interaction of causal factors (Fig.~\ref{fig:HTensorFaces}). 
The composite person signature was computed for every test image by employing the multilinear projection~ algoritm\cite{Vasilescu2011,Vasilescu07a}, and signatures were compared with
weighted nearest neighbhor.

To validate the effectiveness of our system on real-world images, we
report results on ``LFW'' dataset
(\emph{LFW}) ~\cite{Huang2007}. This dataset contains 13,233 facial
images of 5,749 people. %(Fig.~\ref{fig:lfwexamples}).
The photos are
unconstrained (\ie ``in the wild''), and include variation due to
pose, illumination, expression, and occlusion. The dataset consists of
10 train/test splits of the data. We report the mean accuracy and
standard deviation across all splits in Table~\ref{tab:results}. Figure~\ref{fig:HTensorFaces}(b-c) depicts the experimental ROC curves.
We follow the supervised \emph{``Unrestricted,
labeled outside data''} paradigm. %Results may be seen in Table\ref{tab:results} 
\\
\vskip-.09in

%by collecting $\sim\hskip-.05in1000$ images 
\noindent
\underline{Results:} While we cannot celebrate closing the gap on human performance, our results are promising.  DeepFace, a CNN model, improved the prior art verification rates on LFW from $70\%$ to $97.35\%$,  %and celebrated closing the gap in human performance, 
%when they employed
by training on $4.4M$ images of $200\times200$ pixels from $4,030$ people, the same order of magnitude as the number of people in the LFW database. 
We trained on less than one percent ($1\%$) of the $4.4$M total images used to train DeepFace.  
%Images were synthetically rendered by Maya %synthetically generated 
Images were rendered from 3D scans of 100 subjects 
%subjects~\cite{Blanz99}, 
with an 
the intraocular distance of approximately 20 pixels and with a facial region captured by $10,414$ pixels (image size $\approx 100 \times 100$ pixels).
We have currently achieved verification rates just shy of $80\%$ on LFW. % using synthetic training data from $100$ people. 
 When data is limited, CNN models do not convergence or generalize.  

%\begin{figure*}[!tbp]
%\centerline{
%\hfill
%%\subfloat[University of Freiburg 3D Morphable Models]
%{\includegraphics[width=.4\linewidth]{figs/rocfr/fr_rocoverlay.png}\label{Label%1}}
%^\hfill
%%\subfloat[LFW]
%{\includegraphics[width=.4\linewidth]{figs/roclfw/lfw_rocoverlay.png}\label{Label2}}
%\hfill}
%\caption[ROC curves on two facial image datasets]{
%(Left) ROC curves for the University of Freiburg 3D Morphable Faces dataset.
%(Right) ROC curves for the LFW dataset.
%The average accuracies are listed next to each method, along with the area under the curve (\emph{AUC}).
%\emph{Parts} refers to using part-based TensorFaces models to separately analyze facial parts.
%\emph{Gaussian}, \emph{Laplacian} refers to using part-based multiresolution TensorFaces models with a Gaussian/Laplacian pyramid to analyze facial parts at different scales.
%\emph{``Weighted''} refers to using a weighted composite signature (Section~\ref{sec:weights}).}
%EKNotes: Should overlay ROC curves for: PCA, GlobalTfaces, RegionTfaces, LaplacianTfaces, Region+LaplacianTfaces. All with wdiag, all with same dimred settings: [100 5 6]. Legend entries should list accuracy.
%\label{fig:rocs}
%\end{figure*}
%\newpage

\vspace{-.04in}
\section{Conclusion}
\label{sec:Conclusion}
\vspace{-.055in}
%In this paper, we propose a unifying tensor model of wholes and parts
In analogy to autoencoders which are inefficient neural network implementation of principal component analysis, a pattern analysis method based in linear algebra, CNNs are neural network implementations of tensor factorizations.  This paper contributes to the tensor algebraic paradigm and models cause-and-effect as multilinear tensor interaction between intrinsic and extrinsic hierarchical causal factors of data formation.  The data tensor is re-conceptualized into a hierarchical data tensor; a unified tensor model of wholes and parts is proposed; and a new compositional hierarchical tensor factorization is derived. 
 Resulting causal factor representations are interpretable, hierarchical, and statistically invariant to all other causal factors. % by employ $2\nd$ order statistics. 
Our approach is demonstrated in the context of facial images by training on a very small set of synthetic images. While we have not closed the gap on human performance, we report encouraging face verification results on two test data sets\textendash the Freiburg, and the Labeled Faces in the Wild datasets. % by training on approximately half of one percent of the number of images needed to train DeepFace.  
 CNN verification rates improved the $70\%$ prior art to $97.35\%$ when they employed $4.4$M images from $4,030$ people, the same order of magnitude as the number of people in the LFW database.  We have currently achieved verification rates just shy of $80\%$ on LFW by employing synthetic images from 100 people for a total of less than one percent ($1\%$) of the total images employed by DeepFace. By comparison, when data is limited, CNN models do not convergence, or generalize.% dataset consisting of real world images.
%In order to achieve a jump of $27\%$ verification rates from $70\%$ to $97.35\%$, 
%To achieve human level performance, DeepFace employed $4.4$M images from $4,039$ people,the same order magnitude as the number of people in the LFW database.

\iffalse
This paper reconceptualizes the data tensor as a hierarchical data tensor, proposes a unified tensor model of wholes and parts, and introduces a new compositional hierarchical tensor factorization. The factorization computes  a hierarchy of object features and represents an object as a compositional representation of wholes and parts that is invariant to extrinsic causal factors that impact image appearance, but hinder recognition. 
We demonstrate our approach in the context of facial images by training on a very small set of synthetically rendered %generated 
images and report encouraging face verification results on two test datasets – the Freiburg dataset, and on the LabeledFace in the Wild (LFW).% dataset consisting of real world images.
\fi
%\iffalse
\renewcommand{\baselinestretch}{.905}%1425}
\vspace{-.045in}
\section*{acknowledgement}
\vspace{-.055in}
Thank you to NYU Prof. Ernie Davis for feedback on this document.
%, and Xiao (Steven) Zheng, Bo-Kun Wang for the experimental efforts.
%\appendix
%\section{Convolution as a Matrix-Vector Multiplication}
%\vspace{-.01in}
%\fi
%\vspace{-.07in}

%\newpage
{\small
\bibliographystyle{ieee}
%\bibliography{egbib}
\vspace{-.05in}
\bibliography{all,ekrefs}
}

%\clearpage
\end{document}